\documentclass[journal]{IEEEtran}
\ifCLASSINFOpdf
\else
\fi

\usepackage{ifpdf}
\usepackage[pdftex]{graphicx}
\usepackage[cmex10]{amsmath}
\usepackage{subfig}
\usepackage{amssymb}
\usepackage{mathrsfs}
\usepackage{amsmath}
\usepackage{indentfirst}
\usepackage{booktabs}
\usepackage{multirow}
\usepackage{graphicx}
\usepackage{epstopdf}
\usepackage{fixltx2e}
\usepackage{soul}
\usepackage{units}
\usepackage{mathtools}
\usepackage{framed}

\usepackage[utf8]{inputenc}
\usepackage{url} % 用于正确显示URL
\usepackage{cite} % 改进引用格式
\usepackage[numbers]{natbib}

\usepackage{caption}
\usepackage{epsfig,graphicx,amssymb,amsmath}
\usepackage{subfig}
\usepackage{diagbox}
\usepackage{color}
\usepackage{setspace}

\usepackage{amsthm}
\usepackage{algorithm}
\usepackage[noend]{algpseudocode}
\algrenewcommand\algorithmicend{\textbf{end}}
\algrenewcommand\algorithmicif{\textbf{if}}
\algrenewcommand\algorithmicfor{\textbf{for}}
\usepackage{booktabs}
\usepackage{diagbox}
\usepackage{float}
\usepackage{epstopdf}
\usepackage{ragged2e}
\renewcommand{\raggedright}{\leftskip=0pt \rightskip=0pt plus 0cm}

\ifCLASSINFOpdf
\else

\fi
\usepackage{xcolor}

\usepackage{xpatch}

\begin{document}
% \title{ARSDU: An Adaptive Routing Scheme Based on Reward Cooperative Distribution and Tracing Mechanism-enabled MARL towards Software-Defined UASNs }
%\title{ARSDU: An Adaptive Routing Scheme Based on Reward Cooperative Distribution and Tracing Mechanism-enabled MARL in Software-Defined UASNs}
% \title{Underwater Autonomous Data Routing in Software-Defined UASNs: A Reward Cooperative Distribution and Tracing Mechanism-enabled MARL Algorithm}
\title{Multi-AUV Ad-hoc network-based Target Tracking: A Value Gradient Guidance Multi-Agent Diffusion Reinforcement Learning Approach}

\author{Jiaao Ma,
Chuan Lin,~\IEEEmembership{Member,~IEEE},
Guangjie Han,~\IEEEmembership{Fellow,~IEEE},
Shengchao~Zhu,~\IEEEmembership{Student Member,~IEEE,}
Qian Zhu,
Ying Liu,
Zhenyu Wang.

\thanks{\emph{Corresponding author: Guangjie Han}} 
%\thanks{Zhenyu Wang and Yuan Liu contributed equally to this research.}
\thanks{Jiaao Ma, Chuan Lin, Qian Zhu, Ying Liu and Zhenyu Wang are with the Software College, Northeastern University, Shenyang, China. (e-mails: 2727746375@qq.com; chuanlin1988@gmail.com; zhuq@swc.neu.edu.cn; liuy@swc.neu.edu.cn;larrywang1019@outlook.com).}
\thanks{Guangjie Han is with the Key Laboratory of Maritime Intelligent Network Information Technology, Ministry of Education, Hohai University, Changzhou, China (e-mail: hanguangjie@gmail.com).}
\thanks{Shengchao Zhu is with the College of Computer Science and Software Engineering, Hohai University, Nanjing, 210013, China (e-mail: zhushengchao77@gmail.com).}
%\thanks{Chuan Lin is with Software College, Northeastern University, Shenyang, China and is also with Key Laboratory of Data Analytics and Optimization for Smart Industry (Northeastern University), Ministry of Education, China (e-mails: ).}

% 	\IEEEcompsocthanksitem{Moshen Guizani is with Department of Computer Science and Engineering, Qatar University, Qatar. E-mail: mguizani@ieee.org.}
 	
}

\markboth{IEEE TRANSACTIONS ON XXXXXX, VOL.~XX, NO.~X, XXX~XXXX}%
{Ma \MakeLowercase{\textit{et al.}}: Multi-AUV Ad-hoc network-based Target Tracking}


\IEEEtitleabstractindextext{%
	\begin{abstract}
Multi-AUV ad-hoc network-based target tracking requires networked autonomous underwater vehicles (AUVs) to cooperatively track maneuvering targets under constrained acoustic communication, dynamic topology, and uncertain ocean disturbances. Although multi-agent reinforcement learning (MARL) enables decentralized coordination through centralized training, existing methods suffer from high-dimensional joint state-action modeling, noise-sensitive policy generation, leading to unstable training and degraded tracking. To address these issues, we propose VGG-MADiffRL, a value-gradient-guided multi-agent diffusion RL algorithm, and MDCA, a diffusion-based hierarchical control architecture.
Leveraging underwater mission characteristics, we model sonar detection mechanisms and ocean current disturbances, formulating cooperative tracking for multi-AUV ad-hoc networks as an MDP. The proposed MDCA constitutes a three-tier closed-loop control framework: a global intelligent control layer, a local online training layer, and a physical action execution layer. This structure enables synergistic optimization across task allocation, local decision processes, and execution feedback.
Within MDCA, the local online training layer is the policy learning framework; VGG-MADiffRL builds on diffusion policies and incorporates value gradients to guide action generation in the reverse denoising process, steering the generated actions towards higher expected returns. It employs twin value networks with joint optimization and soft target updates to mitigate overestimation and training oscillations, promoting more stable convergence.
Experimental results show that VGG-MADiffRL consistently achieves faster convergence, higher tracking accuracy, and smoother training dynamics in cooperative tracking scenarios, validating its effectiveness and practical engineering value in dynamic underwater settings.

%关键词
		
\end{abstract}
	
	}
\maketitle
\IEEEdisplaynontitleabstractindextext

\section{Introduction}

\label{sec:introduction}

\IEEEPARstart{T}{}he ocean contains abundant biological and mineral resources and is a critical domain for sustaining human society and advancing deep ocean strategic initiatives~\cite{BAILEY2023105672}. 
As underwater communication technologies~\cite{jmse11010124}, particularly acoustic communication, have advanced, the research focus has shifted from single AUV systems to AUV swarm systems~\cite{6678293}. 
In such systems, AUVs form autonomously organized, distributed, and intelligent Multi-AUV ad-hoc Networks (MAANs)~\cite{8428404} that enable collaborative execution of complex underwater missions, such as multiple target tracking and encirclement~\cite{LI2026104033, YAN2025121370}, even in GPS-denied or otherwise unavailable environments.

However, underwater environments are dynamic and uncertain~\cite{JIANG2024118498, LUVISUTTO2025127256}. 
Conventional data-transmission-focused architectures treat the network as a passive conduit, making operational consistency difficult given acoustic channel challenges: rapidly shifting topologies and severely constrained bandwidth~\cite{10622437, 11111675}. 
Overcoming this bottleneck requires a fundamental change in AUV network management: from passive data reception to active perception and reasoning about the environment~\cite{PENG2026123385}.
Although Centralized Training with Decentralized Execution (CTDE) has been widely used to address nonstationarity in multi-agent settings~\cite{10.5555/3455716.3455894, jmse13112072}, its policy learning and optimization struggle with precise continuous cooperative tracking, especially when node relationships in the autonomous network change frequently.

In Multi-AUV ad-hoc network cooperative tracking tasks, existing methods based on MARL still face three main challenges:
(1) Insufficient modeling of complex continuous action distributions: traditional deterministic policies struggle to capture the strong interdependencies and couplings of continuous joint actions needed for collaborative decision-making under dynamic topologies, often leading to limited expressiveness and suboptimal decisions~\cite{LUO2024106243};
(2) Training instability caused by relying on a single optimization objective: when policy updates depend solely on a single reward signal or local supervisory signal, they become highly sensitive to intermittent communication failures, which can cause oscillations in policy learning, slow convergence, and performance fluctuations~\cite{Zhang2021, NEURIPS2022_9c1535a0};
(3) Lack of value guidance during sampling: in the reverse denoising process of diffusion models, the absence of explicit value constraints allows sampled actions to drift away from regions of high expected return and introduces ineffective noise, reducing decision consistency, a problem that is especially harmful in low-bandwidth underwater environments~\cite{pmlr-v162-janner22a}.

This work addresses training instability, restricted policy representation, and suboptimal action sampling in cooperative tracking with multi-AUV ad-hoc networks. 
We build on the generative modeling paradigm of diffusion processes. 
Using a joint optimization scheme that combines loss functions informed by value estimates with policy gradients, we propose the Value Gradient Guided Multi-Agent Diffusion Reinforcement Learning (VGG-MADiffRL) algorithm. 
By replacing deterministic actors with diffusion policies and steering the reverse denoising trajectory via value gradients, this framework allows multi-AUV networks to achieve robust policy optimization, faster convergence, and precise cooperative actions under shifting topologies. 
Its core components are as follows.
(1) A diffusion policy architecture that improves the modeling of complex, interdependent cooperative actions.
(2) A dual-objective framework that combines value signals with policy gradients to stabilize training when the topology changes.
(3) A sampling mechanism that uses value gradients to refine denoising trajectories, prioritizing actions with high expected returns.

\begin{enumerate}[]
    \item Framework for policy learning via diffusion models: VGG-MADiffRL replaces the conventional deterministic actor with a diffusion model to construct a generative policy architecture closely integrated with value estimation. This enables stable and robust policy optimization in multi-AUV ad-hoc networks, thereby enhancing the policy capability to model complex, continuous, and highly interdependent action distributions inherent in dynamic cooperative scenarios.

    \item Combined optimization mechanism for actor networks: VGG-MADiffRL incorporates a dual objective optimization scheme that jointly minimizes a loss function informed by value estimates alongside a policy gradient loss. The component guided by value signals leverages global value estimates to constrain the diffusion policy update direction, effectively mitigating policy oscillations caused by single objective optimization under volatile network topologies.

    \item Diffusion sampling mechanism guided by value gradients: VGG-MADiffRL integrates value signals into the reverse denoising process to iteratively refine the denoising trajectory, steering action sampling toward regions of high expected return from the outset. This mechanism reduces the influence of suboptimal or irrelevant actions and enhances training stability in underwater networks with limited bandwidth and autonomously organized topologies.
\end{enumerate}

The remainder of this paper is organized as follows. Section \ref{Section:2} reviews the related work. Section \ref{Section:3} displays the formulation of the problem and the system preliminaries. Section \ref{Section:4} proposes the MDCA framework. Section \ref{Section:5} details the implementation of the multi-AUV cooperative tracking algorithm based on VGG-MADiffRL. Section \ref{Section:6} presents experimental evaluations and results. Finally, Section \ref{Section:7} concludes the paper and outlines future research directions.

\section{Related Works}\label{Section:2}
In this section, we mainly review the latest research works related to the subject, specifically divided into the following two areas: 1) AUV target tracking algorithms based on reinforcement learning/multi-agent reinforcement learning; 2) reinforcement learning methods based on diffusion models.

\subsection{AUV Tracking Algorithm Based on RL}\label{Section:2-1}

In \cite{10636786}, ACL-SAC, a target tracking framework for AUVs that combines an attention-based convolutional LSTM with Soft Actor-Critic (SAC), was proposed. It fuses multi-sensor data through attention-weighted temporal feature extraction and optimizes stochastic policies to balance entropy maximization and reward accumulation. A composite reward function and prioritized experience replay improve training efficiency, enabling robust tracking under ocean current disturbances, sound speed uncertainty, and sparse observations.

In \cite{10.1145/3458380.3459041}, DDPG-SAC was introduced for high-precision path following of underactuated AUVs under ocean currents. The method decouples surge and heading control, uses an enhanced Line-of-Sight (LOS) law to compensate for drift, and applies a low-pass filter to suppress control chattering. By redesigning state/action spaces and the reward function, it handles nonlinear dynamics, time-varying hydrodynamics, and environmental disturbances, achieving strong generalization and stability in simulations.

In \cite{9439903}, AMAML was developed for trajectory tracking under unknown time-varying dynamics. It decomposes the problem into fixed-dynamics subtasks, models tracking as a Markov decision process using LOS guidance, and embeds an attention module to capture hidden dynamic features. Integrating proximal policy optimization with maximum-entropy meta-learning enables fast adaptation across tasks, overcoming poor generalization and model dependency in conventional RL.

Building on these single-agent advances, research has shifted toward multi-agent coordination, where diffusion models are increasingly integrated with multi-agent reinforcement learning to address complex cooperative decision-making in dynamic underwater environments.

\subsection{Diffusion Models-Based MARL}\label{Section:2-2}

In \cite{zhu2024madiff}, MADiff introduces a diffusion-based offline multi-agent reinforcement learning framework that models inter-agent coordination via latent-space attention and unifies decentralized execution with centralized training and teammate modeling. It uses classifier-free guidance to generate high-return trajectories and recovers actions through an inverse dynamics model, effectively addressing extrapolation errors and limited expressivity in offline settings.

In \cite{10.5555/3709347.3743988}, HGCD extends this idea to heterogeneous teams by integrating a heterogeneous graph attention mechanism into the diffusion process, enabling adaptation to unseen team compositions. Combined with offline meta-RL, it achieves policy generalization across teams while supporting decentralized deployment and tackling challenges in data diversity, compositional generalization, and real-world applicability.

In \cite{11193881}, DMADRL applies diffusion models to online decision-making in semantic vehicular edge computing. It formulates denoising as an MDP to jointly optimize semantic task offloading and resource allocation. Using a composite reward (semantic fidelity, priority, energy) and Gumbel-Softmax reparameterization, it handles mixed discrete-continuous action spaces and dynamic communication constraints, improving system utility and latency.

Despite these advances, existing diffusion-based MARL methods are either offline (MADiff, HGCD) or target discrete-continuous hybrid domains like vehicular networks (DMADRL), making them unsuitable for online, purely continuous, fixed-formation multi-AUV tracking in dynamic underwater environments. To bridge this gap, this paper proposes VGG-MADiffRL and the MDCA architecture, which integrate value-gradient-guided diffusion sampling, dual-objective policy optimization, and hierarchical coordination to enhance training stability, action quality, and robustness in complex cooperative underwater tracking tasks.

%全部用
\section{Preliminary Materials}\label{Section:3}
To accurately emulate the obstacle avoidance and target tracking behaviors of multi-AUV ad-hoc networks in dynamic underwater environments, we incorporate ocean current disturbances and inter-node communication constraints into the modeling process, thus capturing realistic underwater network dynamics. Based on this model, the cooperative operation problem of the multi-AUV system is formulated as a Markov decision process (MDP).
%缩写带个空格和前面的描述 

\subsection{Ocean Circulation Modeling}\label{Section:3-1}

Given the uncertain and dynamic underwater environment, sonar is employed for precise relative positioning between AUVs and targets. The AUV emits acoustic waves to scan its surroundings, and a sector coverage receiver array captures echoes from multiple directions, enabling target localization via intensity analysis. The target detection process is modeled using the active sonar equation in Eq.~\eqref{eq1}:
\begin{equation}
    \mathrm{EM} = \mathrm{SL} - 2\mathrm{TL} + \mathrm{TS} - (\mathrm{NL} - \mathrm{DI}) - \mathrm{DT},
    \label{eq1}
\end{equation}
where $\mathrm{EM}$ is the excess margin, $\mathrm{SL}$ is the source level, $\mathrm{TL}$ is the transmission loss, $\mathrm{TS}$ is the target strength, $\mathrm{NL}$ is the ambient noise level, $\mathrm{DI}$ is the directivity index, and $\mathrm{DT}$ is the detection threshold (in dB).

The Navier--Stokes equation is a fundamental governing equation for fluid motion. It accurately characterizes the dynamic behavior of fluids and can be used to analyze and calculate the hydrodynamic forces acting on underwater vehicles in marine environments. The mathematical form of the equation is given in Eq.~\eqref{eq2}:

\begin{equation}
\label{eq2}
\rho \left( \frac{\partial \mathrm{u}}{\partial t} + \mathrm{u} \cdot \nabla \mathrm{u} \right) = -\nabla p + \mu \nabla^2 \mathrm{u} + \mathrm{F},
\end{equation}
where \(\rho\) denotes the density of the fluid, \(\mathrm{u}\) is the velocity field, \(\frac{\partial \mathrm{u}}{\partial t}\) represents the temporal variation of the velocity, \(\mathrm{u} \cdot \nabla \mathrm{u}\) is the convective term, \(\nabla p\) is the pressure gradient, \(\mu\) denotes the dynamic viscosity, \(\nabla^2 \mathrm{u}\) is the diffusion term, and \(\mathrm{F}\) is the external force term.

%第三部分
\subsection{Kalman Filter-Based State Estimation}\label{Section:3-2}

We define the AUV state by its three-dimensional position and velocity, forming a state-space model for Kalman filtering. The state vector is given by Eq.~\eqref{eq3_state}:

\begin{equation}
\label{eq3_state}
x_k = \left[ x_k,\; y_k,\; z_k,\; v_{x,k},\; v_{y,k},\; v_{z,k} \right]^\top,
\end{equation}
where $x_k$, $y_k$, $z_k$ are the position components at step $k$, and $v_{x,k}$, $v_{y,k}$, $v_{z,k}$ the corresponding velocities.

The discrete-time dynamics with control inputs and process noise are described by the state transition model in Eq.~\eqref{eq4_state}:

\begin{equation}
\label{eq4_state}
x_{k+1} = A x_k + B u_k + w_k,
\end{equation}
where $A$ is the state transition matrix, $B$ the control input matrix, $u_k$ the control input, and $w_k$ the process noise.

Measurements are related to the true state through the observation model with additive noise, as given in Eq.~\eqref{eq5_state}:

\begin{equation}
\label{eq5_state}
z_k = H x_k + v_k,
\end{equation}
where $z_k$ is the measurement vector, $H$ the observation matrix, and $v_k$ the observation noise.

A Kalman filter is used to suppress underwater noise and improve the accuracy of the state estimate. In the prediction step, the prior estimate and its covariance are computed from the previous posterior according to Eq.~\eqref{eq6_state}:

\begin{equation}
\label{eq6_state}
\left\{
\begin{aligned}
x_{k|k-1} &= A x_{k-1|k-1} + B u_{k-1}, \\
P_{k|k-1} &= A P_{k-1|k-1} A^\top + Q,
\end{aligned}
\right.
\end{equation}
where $x_{k|k-1}$ and $P_{k|k-1}$ are the predicted state and covariance, and $Q$ is the process noise covariance.

%截止
In the measurement update, the Kalman gain $K_k$ is computed from the prior error covariance and the observation noise covariance, balancing the relative confidence in the model prediction and the measurement, as given in Eq.~\eqref{eq7_state}:

\begin{equation}
\label{eq7_state}
K_k = P_{k|k-1} H^\top \left( H P_{k|k-1} H^\top + R \right)^{-1},
\end{equation}
where $P_{k|k-1}$ is the prior estimation error covariance matrix, $H$ the observation matrix, and $R$ the observation noise covariance matrix.

The prior state estimate is then updated by the observation innovation, which combines the model prediction with the real-time measurement to form the posterior estimate. This step is described by Eq.~\eqref{eq8_state}:

\begin{equation}
\label{eq8_state}
x_{k|k} = x_{k|k-1} + K_k \left( z_k - H x_{k|k-1} \right),
\end{equation}
where $x_{k|k-1}$ and $x_{k|k}$ are the prior and posterior state estimate vectors, $z_k$ is the observation vector at step $k$, and the term $z_k - H x_{k|k-1}$ is the observation innovation (residual) that is mapped by $K_k$ into the state correction.

Finally, the posterior error covariance matrix is updated to reflect the residual uncertainty after incorporating the measurement and to serve as the starting point for the next prediction cycle, as shown in Eq.~\eqref{eq9_state}:

\begin{equation}
\label{eq9_state}
P_{k|k} = \left( I - K_k H \right) P_{k|k-1},
\end{equation}
where $P_{k|k}$ and $P_{k|k-1}$ are the posterior and prior estimation error covariance matrices, $I$ is the identity matrix, and $K_k H$ represents the combined gain and observation mapping.

%第三部分
\subsection{Markov Process Modeling}\label{Section:3-3}

In the underwater cooperative decision-making process of a multi-AUV system, the interaction with the environment is formalized as a Markov Decision Process (MDP). The MDP is defined by the tuple in Eq.~\eqref{eq10}:
\begin{equation}
\label{eq10}
\mathcal{M} = (\mathcal{S}, \mathcal{A}, \mathcal{P}, \mathcal{R}, \gamma)
\end{equation}
where $\mathcal{S}$ is the state space, $\mathcal{A}$ the action space, $\mathcal{P}$ the state transition probability, $\mathcal{R}$ the reward function, and $\gamma$ the discount factor.

The state space $\mathcal{S}=\{s_1,s_2,\ldots,s_n\}$ contains the states of all AUVs. The state of the $i$-th AUV, $s_i$, combines its ego state $\eta_i$, an environmental perception $\phi_i$, and an observation vector $o_i$. The ego state includes position $p_i\in\mathbb{R}^3$ and velocity $v_i\in\mathbb{R}^3$. The observation vector $o_i=(\kappa_i,\sigma_i,\lambda_i)$ captures the relative position information of targets, neighboring AUVs, and environmental landmarks, respectively. Its dimension depends on the number of targets $N_{\kappa}$, the number of neighbors $N_{\sigma}$, and the number of landmarks $N_{\lambda}$.

The action space $\mathcal{A}=\{a_1,a_2,\ldots,a_n\}$ represents the continuous actions of the multi-AUV system. For AUV $i$, the action vector $a_i\in\mathbb{R}^3$ provides control inputs along the $x$, $y$, and $z$ axes, i.e., $a_i=[a_{x,i},a_{y,i},a_{z,i}]^\top$. Actions are generated by a diffusion policy. To meet execution constraints, the policy outputs undergo range clipping and scaling in the environment execution layer before being converted into executable control commands.

The state transition probability $\mathcal{P}$ captures the dynamic evolution of system states and is built on a physical kinematic model. The discount factor $\gamma\in[0,1)$ balances the weight of future returns. Together, these components define the long-term cumulative return used in policy evaluation.

The reward function $\mathcal{R}$ accounts for several factors: it explicitly incorporates tracking accuracy, collision avoidance, and environmental constraints to improve cooperative multi-AUV tracking performance. The precise formulation is given in Section~\ref{Section:5}.

%以上已经修改完毕

%第四部分
\section{Hierarchical Multi-Agent Collaborative Control Architecture Based on Value Gradient-Guided Diffusion Strategy MARL}\label{Section:4}

In this section, we present the proposed multi-agent diffusion-based collaborative architecture (MDCA), a CTDE reinforcement learning framework specifically designed to address the highly dynamic topology and communication-constrained nature of multi-AUV ad-hoc networks. Building on this architecture, we introduce the value gradient-guided multi-agent diffusion reinforcement learning algorithm (VGG-MADiffRL).

\subsection{Overview of MDCA}\label{Section:4-1}

This subsection presents a hierarchical collaborative control architecture for multi-agent systems under the centralized training with decentralized execution framework. By decomposing global coordination objectives into local operational domains, the proposed design strengthens cooperative consistency within each cluster while decoupling interactions between distinct regions.

\begin{figure}[bth]
	\centering
	\includegraphics[width=0.9\linewidth]{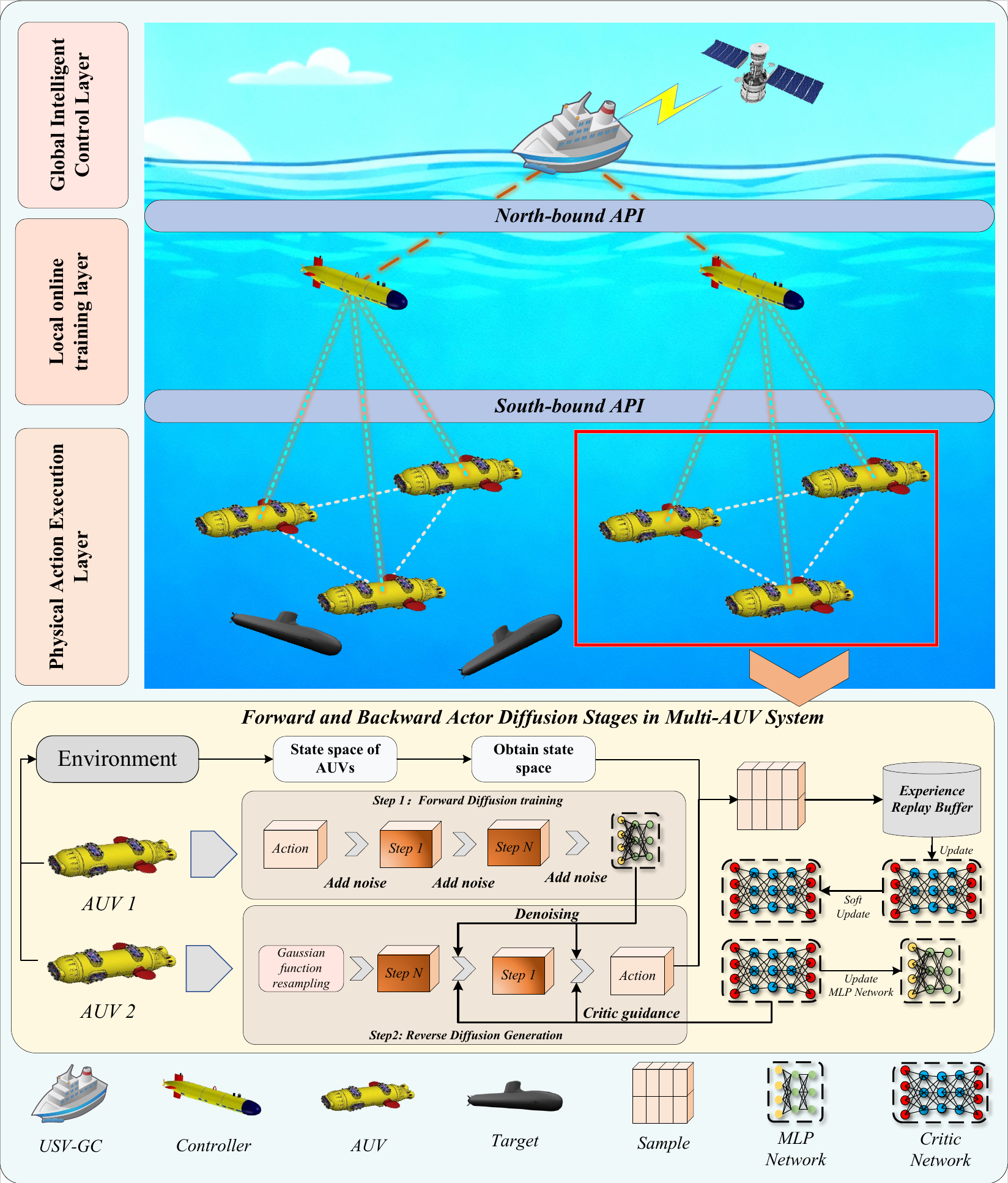}
	\caption{Multi-Agent Diffusion-Based Collaborative Architecture}
	\label{fig1}
\end{figure}

As illustrated in Fig.~\ref{fig1}, the collaborative control architecture for multi-AUV ad-hoc networks comprises three layers: the global intelligent control layer, the local online training layer, and the physical action execution layer. This design aligns well with ad-hoc network characteristics: decentralization, self-organization, dynamic topology, and distributed collaboration. The functional principles and implementation details of these layers are described in the following subsections.

\textbf{Global Intelligent Control Layer:}
The global intelligent control layer, centered on the Unmanned Surface Vessel-based cooperative gateway (USV-CG), serves as the global coordination entry point of the multi-AUV ad-hoc network. It receives global task commands from the satellite. Through the northbound interface with the underwater controller, this layer performs information parsing, forwarding, and scheduling via a hierarchical block mechanism, enabling efficient interaction with the lower-level local online training layer. The global intelligent control layer also receives the ad-hoc network topology, node states, and task progress reported by the local online training layer in real time, dynamically maintaining the global network topology and jointly performing global task decomposition and policy scheduling based on the mission scenario, link quality, and node resources. This design ensures the multi-AUV ad-hoc network can still stably execute cooperative tracking tasks under topology fluctuations.

\textbf{Local Online Training Layer:}
The local online training layer formulates cooperative decisions for dynamically formed AUV clusters in designated operational domains. It maintains a reliable data link with the global intelligent control layer via the northbound interface to acquire and interpret global mission directives, then validate commands and encapsulate local protocols. Leveraging current node distribution, link connectivity, and available resource margins within its assigned domain, this layer routes mission instructions to designated AUV units through the southbound interface, enabling distributed cooperative execution. Concurrently, it acquires navigation states, channel quality indicators, and execution deviations from the physical actuation layer to stabilize the local network topology and optimize policies online. The layer transmits aggregated local states and mission outcomes to the global tier, forming a continuous feedback loop supplying precise operational data for strategic planning.

\textbf{Physical Action Execution Layer:}
As the terminal layer, the physical action execution tier comprises the AUV swarm for mission execution and perception. Via the southbound interface, it receives trajectory planning and cooperative constraint directives from the local online training layer. Each AUV generates control actions using an Actor diffusion policy: the forward process injects noise to characterize environmental and channel uncertainties, while the reverse process reconstructs feasible control sequences guided by critic value gradients. These actions satisfy cooperative constraints and enable high-precision target tracking. Operational data (navigation states and topological variations) is uploaded to an experience replay buffer. This feedback drives online policy updates in the local training layer, ensuring robust cooperative execution under dynamic conditions.

In summary, the proposed hierarchical architecture operates under the centralized training with decentralized execution (CTDE) framework and is suitable for Multi-AUV ad-hoc networks. Decomposing global coordination into localized domains, the design strengthens cooperative consistency within each cluster while decoupling interactions between distinct regions. This structure enhances target tracking performance and system robustness under low bandwidth and highly dynamic underwater conditions.

\subsection{Proposed Multi-Agent Reinforcement Learning Algorithm}\label{Section:4-2}

To address training instability and convergence difficulties in multi-AUV cooperative continuous control tasks caused by non-stationarity in multi-agent environments, Q-value estimation bias, and distributional mismatches in experience replay, this paper proposes a value gradient-guided multi-agent diffusion reinforcement learning algorithm (VGG-MADiffRL).

\begin{figure*}[bth]
	\centering
	\includegraphics[width=1.0\linewidth]{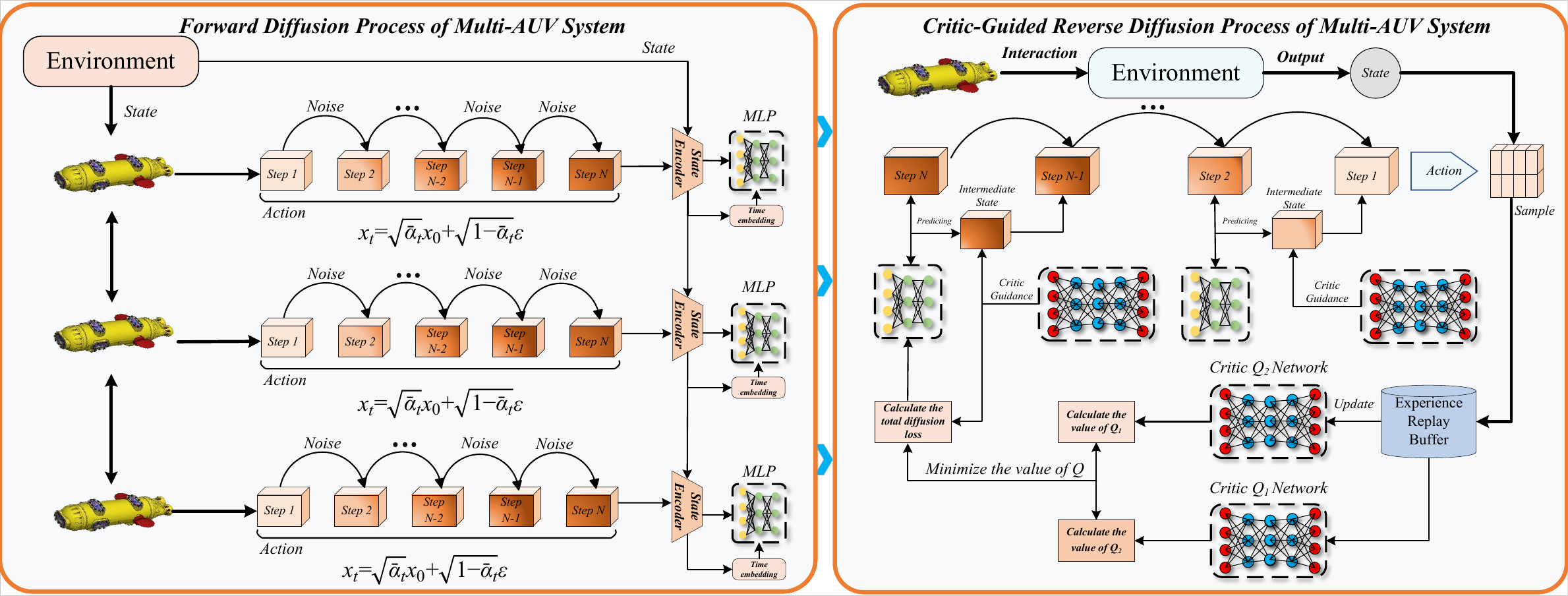}
	\caption{Forward and Critic-Guided Reverse Diffusion Processes of VGG-MADiffRL}
	\label{fig2}
\end{figure*}

\subsubsection{Diffusion Architecture Guided by Value Gradients}\label{Section:4-2-1}

Under the constraints of multi-AUV ad-hoc networks, the forward and reverse diffusion processes of the VGG-MADiffRL algorithm are illustrated in Fig.~\ref{fig2}. The framework explicitly accounts for key underwater characteristics of ad-hoc networks—namely, limited bandwidth, highly dynamic topology, and high communication latency. During the forward diffusion phase, progressive noise injection is applied to construct state-action sequences that emulate the dual uncertainties arising from both the complex marine environment and volatile ad-hoc network conditions. In the reverse diffusion phase, a multilayer perceptron (MLP) network performs iterative denoising guided by value gradients, enabling accurate reconstruction of the original action sequence that satisfies cooperative tracking requirements—all under distributed execution with minimal communication overhead.

\textbf{Forward Diffusion Process:} The forward diffusion process defines how noise is gradually added to the initial action, enabling the marginal distribution and sampling form of the noisy action at any diffusion step to be computed in closed form. The noise schedule coefficient and cumulative signal retention are defined in Eq.~\eqref{eq11}:
\begin{equation}
\label{eq11}
\alpha_t = 1-\beta_t, \qquad \bar{\alpha}_t = \prod_{i=1}^{t} \alpha_i,
\end{equation}
where $\beta_t$ is the noise variance coefficient in step $t$, $\alpha_t$ denotes the signal retention ratio per step, and $\bar{\alpha}_t$ denotes the cumulative signal retention ratio from the initial timestep to step $t$.

Given the initial action sample, the corresponding marginal distribution is described in Eq.~\eqref{eq12}:
\begin{equation}
\label{eq12}
q(\mathrm{x}_t \mid \mathrm{x}_0)
= \mathcal{N}\!\bigl(
\mathrm{x}_t;\,
\sqrt{\bar{\alpha}_t}\,\mathrm{x}_0,\,
(1-\bar{\alpha}_t)\mathrm{I}
\bigr),
\end{equation}
where $\mathcal{N}$ denotes a Gaussian distribution, the mean term $\sqrt{\bar{\alpha}_t}\,\mathrm{x}_0$ represents the scaled initial action, and the covariance term $(1-\bar{\alpha}_t)\mathrm{I}$ represents the accumulated noise covariance over time.

The sampling form based on the reparameterization trick is given in Eq.~\eqref{eq13}:
\begin{equation}
\label{eq13}
\mathrm{x}_t = \sqrt{\bar{\alpha}_t}\,\mathrm{x}_0 + \sqrt{1-\bar{\alpha}_t}\,\boldsymbol{\epsilon},
\qquad
\boldsymbol{\epsilon} \sim \mathcal{N}(\mathbf{0}, \mathrm{I}),
\end{equation}
where $\mathrm{x}_t$ is the noisy action sample at step $t$, and $\boldsymbol{\epsilon}$ is Gaussian noise drawn from the standard normal distribution. This formulation enables direct sampling from the initial action to any diffusion step through reparameterization.

\textbf{Reverse Diffusion Process Guided by Value Gradients:} The reverse sampling process progressively reconstructs the initial action from noisy actions through denoising, while further improving action quality with value-function-guided gradients. The reconstruction estimate of the clean action is computed in Eq.~\eqref{eq14}:
\begin{equation}
\label{eq14}
\hat{\mathrm{x}}_0
= \frac{1}{\sqrt{\bar{\alpha}_t}}
\left(
\mathrm{x}_t
- \sqrt{1-\bar{\alpha}_t}\,
\boldsymbol{\epsilon}_\theta(\mathrm{x}_t, t, \mathrm{s}_t)
\right),
\end{equation}
where $\hat{\mathrm{x}}_0$ is the denoised action estimate reconstructed from the reverse process, and $\boldsymbol{\epsilon}_\theta(\mathrm{x}_t, t, \mathrm{s}_t)$ is the output of the noise prediction network at timestep $t$, which is used to recover the initial action estimate from the noisy action $\mathrm{x}_t$.

The value-function-guided correction is given in Eq.~\eqref{eq15}:
\begin{equation}
\label{eq15}
\hat{\mathrm{x}}_0 = \hat{\mathrm{x}}_0 + \lambda\,\nabla_{\hat{\mathrm{x}}_0} Q_\phi(\mathrm{s}, \hat{\mathrm{x}}_0),
\end{equation}
where $\hat{\mathrm{x}}_0$ is the action estimate after value gradient guidance, $Q_\phi(\mathrm{s}, \hat{\mathrm{x}}_0)$ is the Critic value function, $\nabla_{\hat{\mathrm{x}}_0} Q_\phi$ denotes the gradient of the value function with respect to the action, pointing toward the direction that most rapidly increases the expected return, and $\lambda$ is the guidance strength coefficient that controls the influence of the value gradient.

The value gradient correction in Eq.~\eqref{eq15} applies only during sampling. 
Gradients through this correction term are truncated, preventing direct parameter updates for the diffusion policy. 
Instead, this guidance is captured by the joint Actor loss in Eq.~\eqref{eq:23}, where $\mathcal{L}_{Q\text{-}guide}$ steers the value-guided action estimate $\hat{\mathrm{x}}_0$ toward high-value regions, while $\mathcal{L}_{pg}$ enhances expected returns through differentiable sampled actions. 
This design circumvents explicit higher-order gradient computation from the value signal during Actor optimization, improving training stability.

The reverse diffusion sampling update is completed by Eq.~\eqref{eq16}:
\begin{equation}
\label{eq16}
\mathrm{x}_{t-1}
= \sqrt{\bar{\alpha}_{t-1}}\,\hat{\mathrm{x}}_0
+ \sqrt{1-\bar{\alpha}_{t-1}-\sigma_t^2}\,\boldsymbol{\epsilon}_\theta(\mathrm{x}_t, t, \mathrm{s}_t)
+ \sigma_t \mathbf{z},
\end{equation}
where $\mathrm{x}_{t-1}$ is the action sample at step $t-1$ in the reverse process, the first term $\sqrt{\bar{\alpha}_{t-1}}\,\hat{\mathrm{x}}_0$ is the guided action estimate, the second term is the predicted noise component, and the third term $\sigma_t \mathbf{z}$ is the injected random noise, where $\mathbf{z}$ follows the standard normal distribution. This equation realizes the denoising transition from timestep $t$ to timestep $t-1$.

The noise standard deviation of the reverse process is determined by Eq.~\eqref{eq17}:
\begin{equation}
\label{eq17}
\sigma_t = \eta\sqrt{\frac{1-\bar{\alpha}_{t-1}}{1-\bar{\alpha}_t}}\sqrt{1-\frac{\bar{\alpha}_t}{\bar{\alpha}_{t-1}}},
\end{equation}
where $\sigma_t$ is the noise standard deviation used in reverse sampling at step $t$, and $\eta$ is the randomness control coefficient (DDPM parameter). When $\eta=0$, the sampling is deterministic; when $\eta=1$, the sampling is fully stochastic. This parameter balances the diversity and stability of the generation process.

\textbf{Value Gradient Estimation:} The value gradient is defined in Eq.~\eqref{eq18} to quantify the sensitivity of the Critic value function to action variations:
\begin{equation}
\label{eq18}
\nabla_{\mathrm{a}_0} Q_\phi(\mathrm{s}, \mathrm{a}_0)
= \left.
\begin{bmatrix}
\frac{\partial Q_\phi}{\partial a^{(1)}} & \cdots & \frac{\partial Q_\phi}{\partial a^{(d)}}
\end{bmatrix}^\top
\right|_{\mathrm{a}=\mathrm{a}_0},
\end{equation}
where $\frac{\partial Q_\phi}{\partial a^{(i)}}$ is the partial derivative of the value function with respect to the $i$-th action component, and $(\cdot)^\top$ denotes vector transpose. This gradient vector points toward the direction of the fastest increase in the value function.

\subsubsection{Joint Optimization Loss of VGG-MADiffRL}\label{Section:4-2-2}

In the design of the loss function, we explicitly account for both algorithmic stability and the generative characteristics of the diffusion model, decomposing the overall optimization objective into two components: the dual critic loss and the diffusion-based actor loss.

\textbf{Dual Critic Loss:} The critic loss is constructed based on the temporal difference (TD) error. First, the target Q-value for the $i$-th mini-batch sample is computed as given in Eq.~\eqref{eq:19}, and then the mean squared error is used to measure the deviation between the online critics and the target Q-value, as shown in Eq.~\eqref{eq:20}:
\begin{equation}
    y_i = r_i + \gamma (1 - d_i) \min \left\{ Q'_{1,\phi_i}(s', a'),\, Q'_{2,\phi_i}(s', a') \right\},
    \label{eq:19}
\end{equation}
\begin{equation}
    \mathcal{L}_{\text{critic}}^{(i)} = \left\| Q_{1,\phi_i}(s, a) - y_i \right\|^2 + \left\| Q_{2,\phi_i}(s, a) - y_i \right\|^2,
    \label{eq:20}
\end{equation}
where $s$, $s'$ denote the current and next global states input to the critics; $a$ is the current joint action of all agents; $a'$ is the next joint action generated by the target actor networks; $r_i$ and $d_i$ are the immediate reward and termination flag of agent $i$; $\gamma$ is the reward discount factor; $Q_{1,\phi_i}$, $Q_{2,\phi_i}$ are the dual online critic networks for agent $i$; and $Q'_{1,\phi_i}$, $Q'_{2,\phi_i}$ are the corresponding dual target critic networks. Under the CTDE framework, the centralized critics receive the global state $s$ and joint action $a$, while the diffusion-based actors condition on each agent's local observation $o_i$ to generate actions.

\textbf{Diffusion Actor Loss:} The Q-guided loss term of the diffusion-based actor guides action generation using the Q-values output by the critics, with its mathematical form given in Eq.~\eqref{eq:21}:
\begin{equation}
    \mathcal{L}_{Q\text{-guide}} = -\lambda_q \cdot \frac{1}{N} \sum_{i=1}^{N} \min \left\{ Q_1(s, a_{0,i}),\, Q_2(s, a_{0,i}) \right\},
    \label{eq:21}
\end{equation}
where $\lambda_q$ is the Q-guidance coefficient, and $a_{0,i}$ is the denoised action estimate used in the Q-guidance term.

The policy gradient loss directly maximizes the action value evaluated by the critics, with its mathematical expression given in Eq.~\eqref{eq:22}:
\begin{equation}
    \mathcal{L}_{\text{pg}} = -\frac{1}{N} \sum_{i=1}^{N} \min \left\{ Q_1(s, a_i),\, Q_2(s, a_i) \right\},
    \label{eq:22}
\end{equation}
where $a_i$ is the differentiable sampled action used in the policy gradient term.

The total actor loss combines both components:
\begin{equation}
    \mathcal{L}_{\text{actor}} = \mathcal{L}_{Q\text{-guide}} + \mathcal{L}_{\text{pg}},
    \label{eq:23}
\end{equation}
where $\mathcal{L}_{\text{actor}}$ is the final actor loss, $\mathcal{L}_{Q\text{-guide}}$ is the Q-guided loss defined in Eq.~\eqref{eq:21}, and $\mathcal{L}_{\text{pg}}$ is the policy gradient loss defined in Eq.~\eqref{eq:22}.

\section{Proposed Multi-AUV Cooperative Target Tracking Scheme}\label{Section:5}
This paper considers cooperative target tracking in multi-AUV ad-hoc networks. Using the proposed VGG-MADiffRL algorithm, we present the key tracking strategy for ad-hoc network constraints and describe the algorithm's complete execution pipeline.

\subsection{Proposed Reward Function for Multi-AUV Tracking}\label{Section:5-1}
We present a composite reward function for multi-AUV ad-hoc networks. In the RL framework, cooperative tracking under ad-hoc network constraints is a Markov decision process (MDP) maximizing cumulative reward, guiding AUVs toward stable coordination in bandwidth-limited, dynamic, and communication-impaired underwater environments. We design three reward components for tracking fidelity, inter-agent safety, and environmental constraints, ensuring efficiency and long-term policy stability in complex ad-hoc network scenarios.

The total reward for the $i$-th AUV at the current timestep is given by Eq.~\eqref{eq:24}:
\begin{equation} 
    R_i = \alpha  \cdot r_{\text{pos}} + \beta \cdot r_{\text{col}} + \delta \cdot r_{\text{land}},
    \label{eq:24}
\end{equation}
where $R_i$ denotes the instantaneous scalar reward received by the $i$-th AUV; $\alpha$, $\beta$, and $\delta$ are positive weighting coefficients that balance the influence of the target position reward ($r_{\text{pos}}$), the collision penalty ($r_{\text{col}}$), and the landmark constraint term ($r_{\text{land}}$) on policy learning.

To emulate realistic underwater tracking environments, the target position reward is defined as given in Eq.~\eqref{eq:25}:
\begin{equation}
    r_{\text{pos}} = \begin{cases}
        -d_t, & d_t > d_{t,\min}, \\
        w d_t - (w+1) d_{t,\min}, & d_t \leq d_{t,\min},
    \end{cases}
    \label{eq:25}
\end{equation}
where $d_t$ denotes the distance between the agent and the target, $d_{t,\min}$ is the threshold distance defining the target's proximity zone, and $w$ is a reward modulation parameter within this zone.

To prevent collisions among nodes during dense cooperative maneuvers in the ad-hoc network, a safety constraint penalty is designed based on relative inter-agent distances. The collision penalty is formulated as shown in Eq.~\eqref{eq:26}:
\begin{equation}
    r_{\text{col}} = \begin{cases}
        -\lambda_1 (d_{o,\min} - d_o)^2, & d_o < d_{o,\min}, \\
        -\min(\lambda_2, \lambda_3 d_o), & d_o \geq d_{o,\min},
    \end{cases}
    \label{eq:26}
\end{equation}
where $d_o$ represents the relative distance between two AUVs, $d_{o,\min}$ is the minimum allowable safe distance, $\lambda_1$ controls the penalty intensity for close-range proximity, $\lambda_2$ sets the upper bound for long-range penalties, and $\lambda_3$ is the linear slope coefficient governing the penalty decay.

To smoothly model landmark region constraints within the reward function, a Sigmoid function is introduced, as expressed in Eq.~\eqref{eq:27}:
\begin{equation}
    \sigma(x) = \frac{1}{1 + e^{-x}},
    \label{eq:27}
\end{equation}
where $\sigma(x)$ serves as a smooth activation function that transforms hard-threshold penalty relationships into a continuously differentiable form, thereby enhancing training stability.

Building upon Eq.~\eqref{eq:27}, the obstacle avoidance penalty is defined as:
\begin{equation}
    r_{\text{land}} = -\lambda_l \sum_{k=1}^{3} \sigma\left( \frac{\tau_k - d_{l,k}}{s_k} \right),
    \label{eq:28}
\end{equation}
where $\lambda_l$ is the landmark penalty weight, $d_{l,k}$ denotes the distance between the agent and the $k$-th landmark, $\tau_k$ represents the constraint threshold for the corresponding landmark, and $s_k$ is a smoothing adjustment parameter.

In summary, the proposed composite reward function comprises three core components: target tracking accuracy, swarm safety constraints, and underwater obstacle avoidance penalties. This formulation comprehensively captures the primary objectives and operational constraints of multi-AUV ad-hoc networks in dynamic underwater scenarios. By closely mirroring real-world underwater task dynamics, this reward mechanism significantly improves simulation fidelity and effectively enhances the learning efficiency, convergence stability, and cooperative robustness of reinforcement learning models under bandwidth-limited and topologically volatile network conditions.

\subsection{Proposed Tracking Algorithm Based on VGG-MADiffRL}\label{Section:5-2}

\begin{algorithm}[t]
\caption{Proposed Multi-AUV Cooperative Target Tracking Algorithm Based on VGG-MADiffRL}
\label{alg:vgg_madiffrl_tracking}
\begin{algorithmic}[1]
\Require Number of episodes $N_{\text{ep}}$, episode length $L_{\text{ep}}$, batch size $B$, update interval $I_{\text{update}}$, minimal buffer size $S_{\min}$, soft update coefficient $\tau$, guidance interval $I_{\text{guide}}$
\Ensure Trained diffusion policies and twin critics for multi-AUV cooperative tracking
\State Initialize environment $\mathit{env}$ and VGG-MADiffRL model
\State Initialize replay buffer $\mathcal{D}$
\State Set global step counter $t_{\text{total}} \gets 0$
\For{$e = 1$ \textbf{to} $N_{\text{ep}}$}
    \State Reset environment and obtain initial state $\mathbf{s}$
    \For{$t = 1$ \textbf{to} $L_{\text{ep}}$}
        \State Set critic-guidance flag $g_t \gets \mathbb{I}(t \bmod I_{\text{guide}} = 0)$
        \For{each agent $i = 1$ \textbf{to} $N$}
            \If{$g_t = 1$}
                \State Generate action $\mathbf{a}_i$ by guided diffusion sampling (Eqs.~\eqref{eq14}--\eqref{eq18})
            \Else
                \State Generate action $\mathbf{a}_i$ by diffusion policy without guidance
            \EndIf
        \EndFor
        \State Execute joint action $\mathbf{a} = (\mathbf{a}_1,\ldots,\mathbf{a}_N)$ in $\mathit{env}$
        \State Observe next state $\mathbf{s}'$, reward $\mathbf{r}$, and done flag $\mathbf{d}$
        \State Store transition $(\mathbf{s},\mathbf{a},\mathbf{r},\mathbf{s}',\mathbf{d},0)$ in $\mathcal{D}$
        \State Update current state $\mathbf{s} \gets \mathbf{s}'$
        \State $t_{\text{total}} \gets t_{\text{total}} + 1$

        \If{$|\mathcal{D}| \ge S_{\min}$ \textbf{and} $t_{\text{total}} \bmod I_{\text{update}} = 0$}
            \State Sample minibatch $\mathcal{B}$ from $\mathcal{D}$
            \For{each agent $i = 1$ \textbf{to} $N$}
                \State Construct target joint action $\mathbf{a}'$ from target diffusion policies
                \State Compute target value $y_i$ using Eq.~\eqref{eq:19}
                \State Update twin critics by minimizing $\mathcal{L}_{\text{critic}}^{(i)}$ in Eq.~\eqref{eq:20}
                \State Update diffusion actor by minimizing $\mathcal{L}_{\text{actor}}^{(i)}$ in Eq.~\eqref{eq:23}
                \State \hspace{1em} where $\mathcal{L}_{Q\text{-guide}}^{(i)}$ and $\mathcal{L}_{pg}^{(i)}$ are defined in Eqs.~\eqref{eq:21}--\eqref{eq:22}
            \EndFor
            \State Soft update Actor and Critic target networks for all agents by $\theta_{\text{target}} \gets (1-\tau)\theta_{\text{target}} + \tau\theta$.
        \EndIf

        \If{$\mathbf{d} = \text{True}$}
            \State \textbf{break}
        \EndIf
    \EndFor
\EndFor
\end{algorithmic}
\end{algorithm}

\begin{figure*}[bth]
	\centering
	\includegraphics[width=1.0\linewidth]{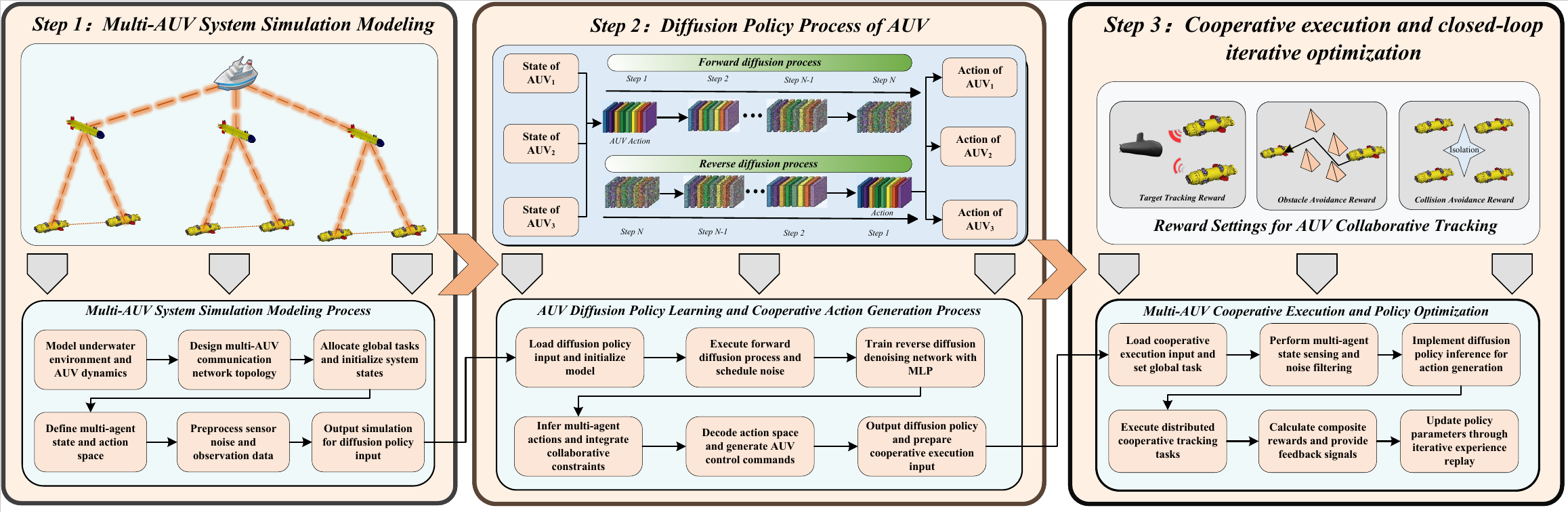}
	\caption{Workflow of the Generative Multi-AUV MARL Algorithm}
	\label{fig3}
\end{figure*}

In this section, the proposed VGG-MADiffRL-based tracking algorithm is formalized in Algorithm~\ref{alg:vgg_madiffrl_tracking}. The complete workflow for ad-hoc networks is decomposed into three steps, as illustrated in Fig.~\ref{fig3}.

\textbf{Step 1: Simulation Environment Construction.} The underwater simulation framework for multi-AUV ad-hoc networks integrates 3D environmental modeling, acoustic channel simulation, and sonar-based state representation, capturing hydrodynamic and topological uncertainties within an MDP and providing a robust validation platform.

\textbf{Step 2: Diffusion Policy Architecture Design.} A diffusion-based policy learning framework is deployed where the forward process injects progressive noise to emulate environmental and network fluctuations. During the reverse process, Critic value gradients guide iterative denoising to reconstruct high-quality cooperative action sequences, ensuring precise distributed decision-making under bandwidth constraints.

\textbf{Step 3: Closed-Loop Cooperative Execution.} In distributed execution, each AUV acts autonomously on local observations and generates actions via the trained diffusion policy. Interaction trajectories are stored in the experience replay buffer. Periodically, buffered samples are used to update the diffusion policy and dual Critic networks through soft target updates. This closed-loop paradigm enables the multi-AUV system to sustain adaptive coordination and robust target tracking in dynamic underwater environments.

The proposed VGG-MADiffRL-based tracking algorithm appears in Algorithm~\ref{alg:vgg_madiffrl_tracking}, integrating guided diffusion sampling, twin-critic learning, and soft target updates into a unified multi-AUV loop. 
Algorithm~\ref{alg:vgg_madiffrl_tracking} initializes the environment, replay buffer, and global timestep counter $t_{\text{total}}$ (Lines 1--3) and iterates over episodes (Line 4). 
Each episode resets environment (Line 5) and proceeds over timesteps (Line 6). 
At each timestep, $t \bmod I_{\text{guide}}$ determines guidance flag $g_t$ (Line 7), and each agent generates actions by either value-gradient-guided diffusion sampling (Eqs.~\eqref{eq14}--\eqref{eq18}, Lines 9--10) or unguided diffusion policy sampling (Lines 11--12). 
The joint action is executed in the environment, and the next state, reward, and done flag are observed (Lines 13--14). 
The transition $(\mathbf{s},\mathbf{a},\mathbf{r},\mathbf{s}',\mathbf{d},0)$ is stored in replay buffer $\mathcal{D}$, and the current state and global timestep counter are updated (Lines 15--17). 
When $|\mathcal{D}| \ge S_{\min}$ and $t_{\text{total}} \bmod I_{\text{update}} = 0$ (Line 18), a minibatch is sampled from $\mathcal{D}$ (Line 19), and each agent is updated (Line 20): target joint actions constructed from target diffusion policies (Line 21), target values computed via Eq.~\eqref{eq:19} (Line 22), twin critics optimized via Eq.~\eqref{eq:20} (Line 23), and diffusion actors optimized via Eq.~\eqref{eq:23}, with $\mathcal{L}_{Q\text{-guide}}$ and $\mathcal{L}_{pg}$ defined in Eqs.~\eqref{eq:21}--\eqref{eq:22} (Lines 24--25). 
Each update cycle concludes with soft updates of Actor and Critic target networks via $\theta_{\text{target}} \gets (1-\tau)\theta_{\text{target}} + \tau\theta$ (Line 26). 
The episode terminates early if the done flag is true (Lines 27--28). 
This procedure enables stable, sample-efficient cooperative tracking by coupling value-guided diffusion action generation with twin-critic-based policy optimization.

\section{Evaluations}\label{Section:6}
This section presents a comprehensive experimental evaluation of the proposed algorithm. The cooperative tracking performance of the multi-AUV ad-hoc network is analyzed across multiple metrics, including average reward, convergence stability, tracking accuracy, and robustness. Comparative experiments with state-of-the-art multi-agent reinforcement learning algorithms are conducted to validate the effectiveness of VGG-MADiffRL.

\subsection{Simulation Setup}\label{Section:6-1}

All experiments are conducted on a computational platform equipped with an AMD Ryzen 9 8940HX processor, RTX 5060 GPU, and 16 GB RAM. All code is implemented in Python 3.10.

The simulation environment is built on OceanGym~\cite{xue2025oceangymbenchmarkenvironmentunderwater}, a benchmark environment for underwater embodied agents. We adopt its modular agent-environment interface for standardized multi-agent underwater interaction and extend it with custom hydrodynamic effects and acoustic communication constraints to model the physical dynamics and network conditions of multi-AUV ad-hoc networks.

In the evaluations, the target moves at a predefined constant velocity, while AUVs are initially distributed in a circular ring-shaped region approximately 3.5 to 5 km from the target.

To comprehensively evaluate algorithm performance under varying network scales, four distinct tracking scenarios are employed in the assessment: 10 AUVs tracking 3 targets, 8 AUVs tracking 3 targets, 6 AUVs tracking 2 targets, and 4 AUVs tracking 2 targets.

All the parameters in the evaluations are detailed in Table~\ref{table1}.

\subsection{Results and Discussion}\label{Section:6-2}

%这个地方需要稍微分开阐述基线算法和非基线算法

We compare VGG-MADiffRL against two groups of MARL methods. The first group consists of five MARL algorithms for continuous control: MASAC~\cite{9446746}, MAPPO~\cite{NEURIPS2022_9c1535a0}, MAAC~\cite{doi:10.1142/S0218001422520140}, MATD3~\cite{ackermann2019reducingoverestimationbiasmultiagent}, and MADDPG~\cite{10.5555/3295222.3295385}. The second group consists of two MARL methods designed for underwater AUV scenarios, DSBM~\cite{10814089} and MA-A3C~\cite{10621453}.
Our approach is evaluated mainly from the following aspects: 1) convergence speed; 2) tracking accuracy; 3) mean tracking error; 4) error standard deviation; 5) diffusion steps required for strategy generation in our framework; 6) ablation studies validating each component's contribution to system performance, and 7) system availability under dynamic ad-hoc network conditions.

\begin{table}[H]
    \centering
    \caption{Simulation Parameters}
    \label{table1}
    \begin{tabular}{ccc}
    \hline
    \textbf{Parameter} & \textbf{Description} & \textbf{Value} \\ \hline
    $N_A$ & Number of AUVs & [4,6,8,10] \\
    $N_T$ & Number of Targets & [2,3] \\
    $l_r$ & Learning rate & $1e^{-3}$ \\
    $N_E$ & Training rounds & 4000 \\
    $N_h$ & Hidden layer neurons & 256 \\
    $\gamma$ & Discount factor & 0.95 \\
    $\tau$ & Network update coefficient & $1e^{-2}$ \\
    $d_{\min}^{\phi}$ & Minimum tracking distance & 80 m \\
    $d_{\min}^{\kappa}$ & Minimum AUV distance & 80 m \\
    $L_e$ & Episode length & 400 \\
    $B_s$ & Buffer size & 100,000 \\
    $U_i$ & Update interval & 400 \\
    $M_s$ & Minimal buffer size & 4000 \\
    $B$ & Batch size & 256 \\
    $\rho$ & Fluid density & 1000 kg/m$^3$ \\
    $\mu$ & Fluid viscosity & $10^{-3}$ Pa$\cdot$s \\
    $D$ & Damping factor & 0.25 \\
    $\Delta t$ & Simulation time step & 0.1 s \\
    \hline
\end{tabular}
\end{table}

\begin{figure*}[t!]
    \centering
    % First row
    \subfloat[Scenario of 4 AUVs Tracking 2 Targets]
    {\includegraphics[width=0.47\textwidth]{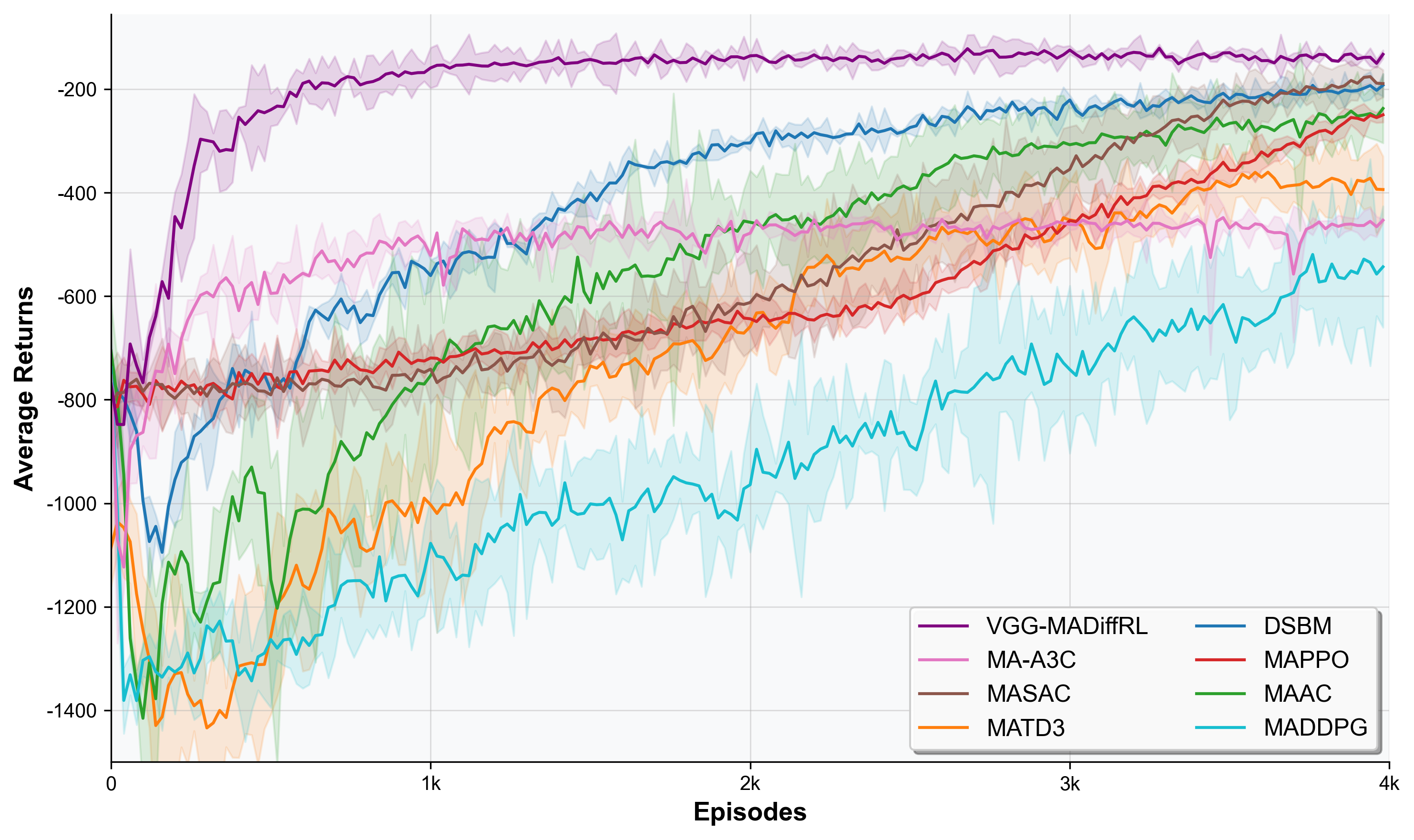}}\hfill
    \subfloat[Scenario of 6 AUVs Tracking 2 Targets]
    {\includegraphics[width=0.47\textwidth]{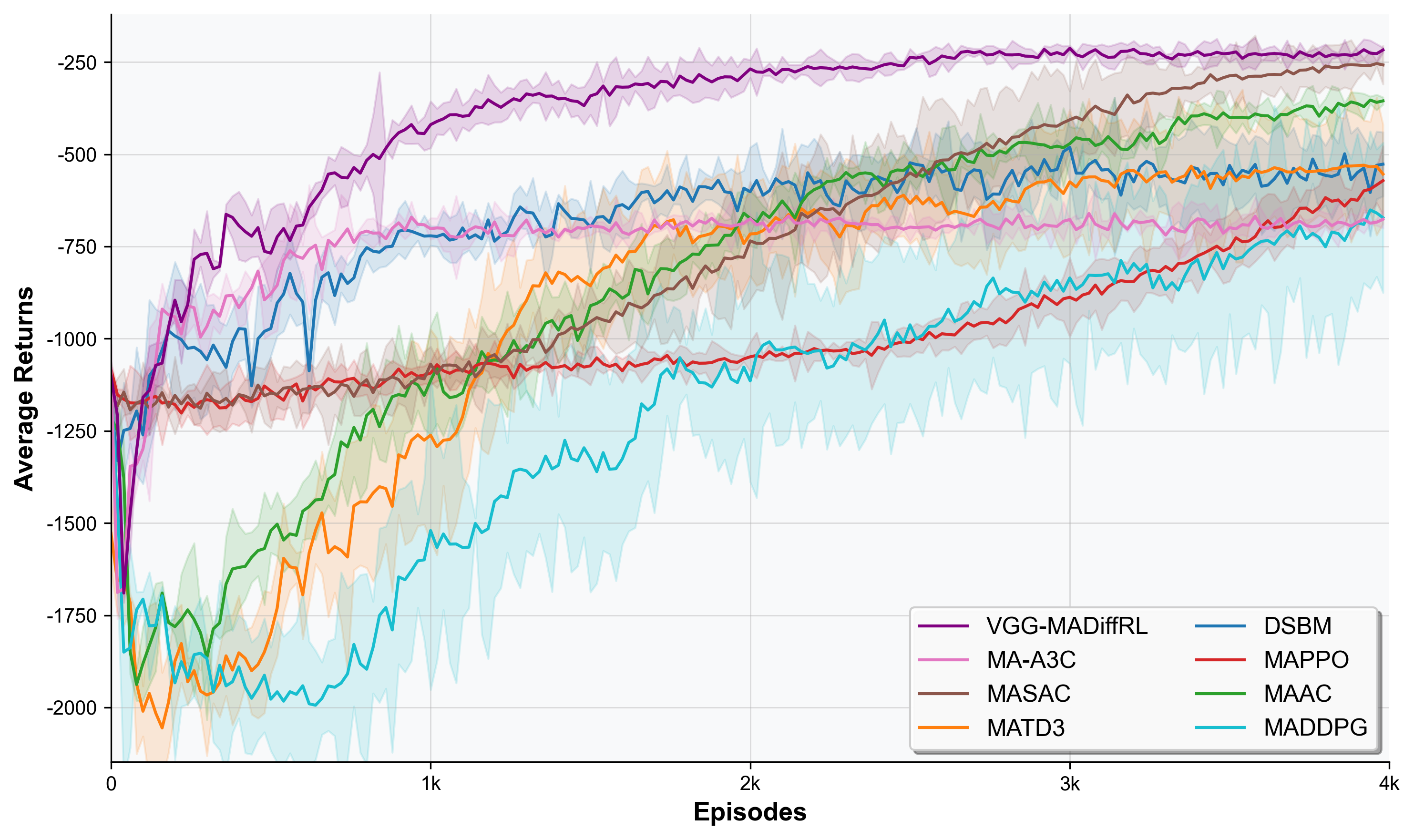}}
    
    \vspace{0.5em} % Vertical space between rows
    
    % Second row
    \subfloat[Scenario of 8 AUVs Tracking 3 Targets]
    {\includegraphics[width=0.47\textwidth]{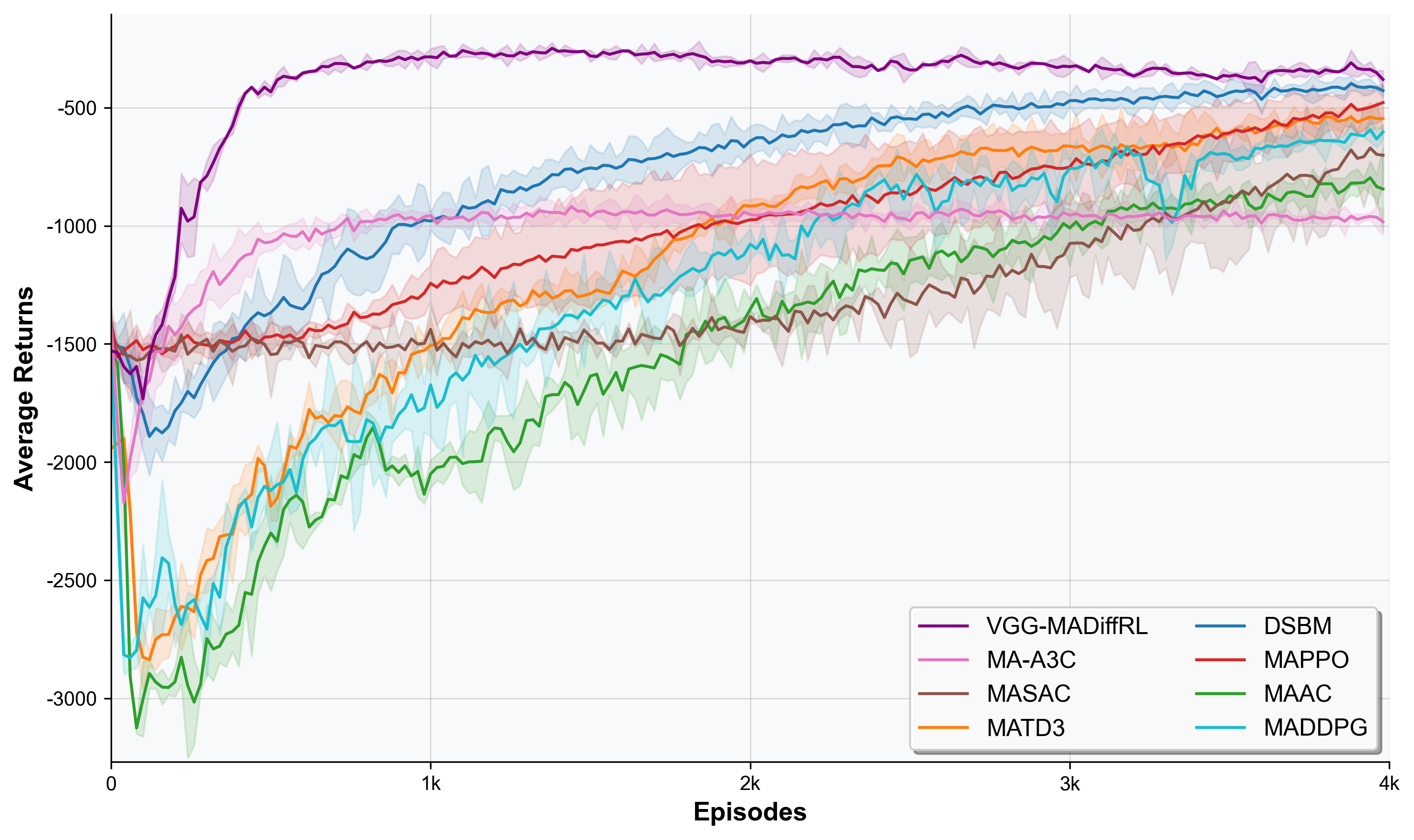}}\hfill
    \subfloat[Scenario of 10 AUVs Tracking 3 Targets]
    {\includegraphics[width=0.47\textwidth]{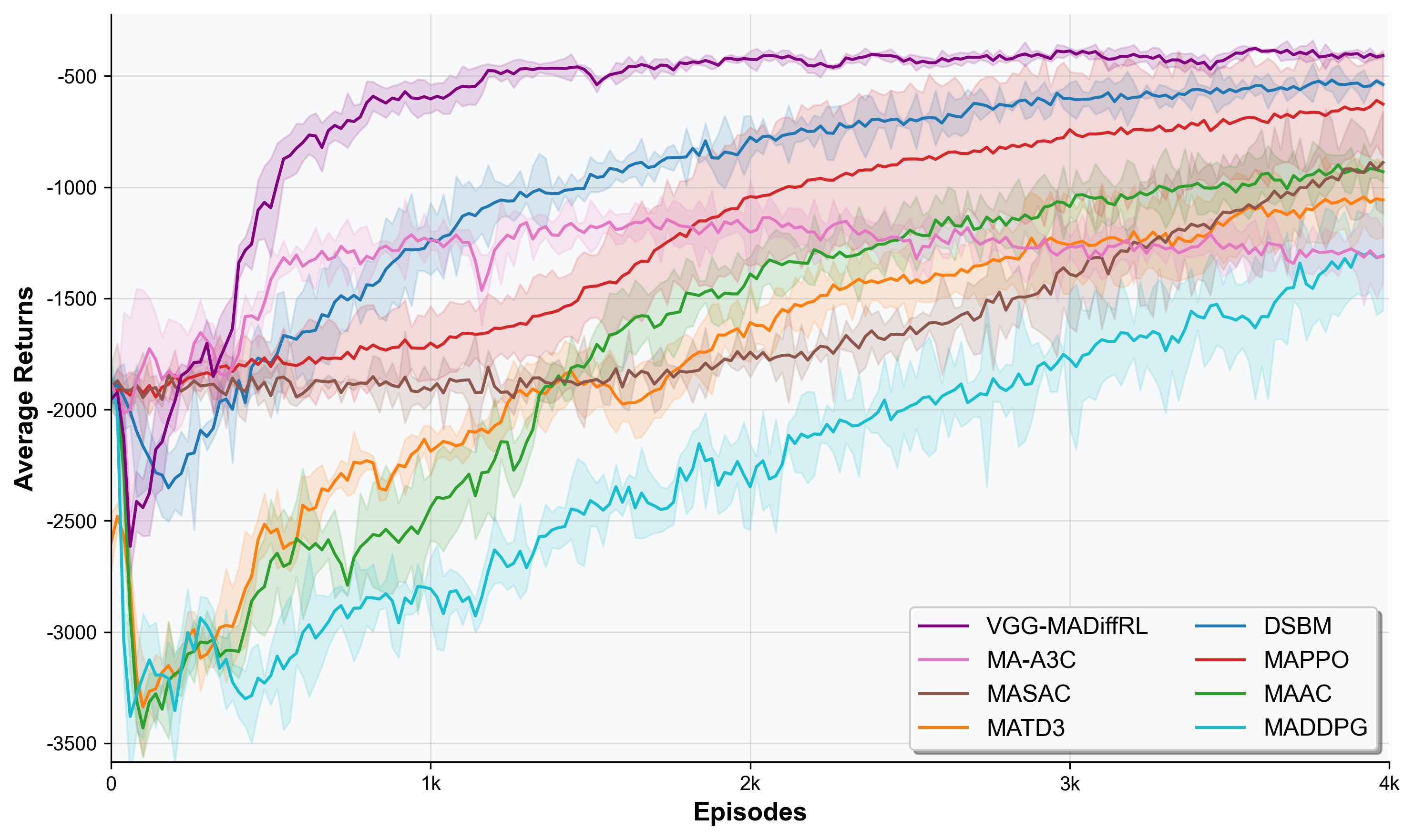}}
    
    \caption{Convergence Speed Evaluation}
    \label{fig6}
\end{figure*}

\begin{table*}[t]
\centering
\caption{Tracking Accuracy Comparison}
\label{tab:tracking_accuracy}
\resizebox{\textwidth}{!}{%
\begin{tabular}{lcccc}
\hline
\multirow{2}{*}{\textbf{Algorithms}} & \textbf{4 AUVs Tracking 2 Targets} & \textbf{6 AUVs Tracking 2 Targets} & \textbf{8 AUVs Tracking 3 Targets} & \textbf{10 AUVs Tracking 3 Targets} \\
\cline{2-5}
 & $Mean \pm SD$ & $Mean \pm SD$ & $Mean \pm SD$ & $Mean \pm SD$ \\
\hline
VGG-MADiffRL & 74.63\%$\pm$0.19\% & 72.73\%$\pm$0.04\% & 77.40\%$\pm$0.05\% & 61.52\%$\pm$0.28\% \\
DSBM & 55.83\%$\pm$0.47\% & 43.83\%$\pm$2.78\% & 38.58\%$\pm$1.86\% & 35.21\%$\pm$1.50\% \\
MA-A3C & 33.42\%$\pm$0.61\% & 31.68\%$\pm$1.89\% & 25.74\%$\pm$0.20\% & 32.29\%$\pm$0.66\% \\
MAPPO & 60.27\%$\pm$0.13\% & 52.82\%$\pm$0.41\% & 67.75\%$\pm$0.11\% & 48.76\%$\pm$0.02\% \\
MASAC & 64.22\%$\pm$0.05\% & 62.97\%$\pm$0.02\% & 45.11\%$\pm$1.59\% & 10.32\%$\pm$1.00\% \\
MAAC & 61.19\%$\pm$0.26\% & 45.07\%$\pm$0.17\% & 41.35\%$\pm$1.13\% & 40.64\%$\pm$0.73\% \\
MATD3 & 64.96\%$\pm$0.62\% & 35.89\%$\pm$0.48\% & 48.84\%$\pm$0.15\% & 49.38\%$\pm$0.26\% \\
MADDPG & 48.91\%$\pm$1.41\% & 26.25\%$\pm$0.27\% & 50.46\%$\pm$0.26\% & 47.61\%$\pm$0.34\% \\
\hline
\end{tabular}
}
\end{table*}

\begin{table*}[t]
\centering
\caption{Mean Tracking Error Comparison}
\label{tab:mean_tracking_error}
\resizebox{\textwidth}{!}{%
\begin{tabular}{lcccc}
\hline
\multirow{2}{*}{\textbf{Algorithms}} & \textbf{4 AUVs Tracking 2 Targets} & \textbf{6 AUVs Tracking 2 Targets} & \textbf{8 AUVs Tracking 3 Targets} & \textbf{10 AUVs Tracking 3 Targets} \\
\cline{2-5}
 & $Mean \pm SD$ & $Mean \pm SD$ & $Mean \pm SD$ & $Mean \pm SD$ \\
\hline
VGG-MADiffRL & 0.1329$\pm$0.0001 & 0.1159$\pm$0.0001 & 0.1099$\pm$0.0001 & 0.1524$\pm$0.0001 \\
DSBM & 0.1629$\pm$0.0009 & 0.1929$\pm$0.0210 & 0.1852$\pm$0.0048 & 0.1966$\pm$0.0006 \\
MA-A3C & 0.3463$\pm$0.0028 & 0.2053$\pm$0.0019 & 0.3718$\pm$0.0017 & 0.2243$\pm$0.0006 \\
MAPPO & 0.1439$\pm$0.0002 & 0.1497$\pm$0.0004 & 0.1312$\pm$0.0004 & 0.2114$\pm$0.0001 \\
MASAC & 0.1433$\pm$0.0002 & 0.1462$\pm$0.0004 & 0.1807$\pm$0.0011 & 0.3086$\pm$0.0000 \\
MAAC & 0.1379$\pm$0.0003 & 0.1962$\pm$0.0008 & 0.2321$\pm$0.0005 & 0.2792$\pm$0.0004 \\
MATD3 & 0.1472$\pm$0.0020 & 0.2541$\pm$0.0046 & 0.1671$\pm$0.0004 & 0.1782$\pm$0.0004 \\
MADDPG & 0.1678$\pm$0.0009 & 0.2010$\pm$0.0005 & 0.1908$\pm$0.0005 & 0.2246$\pm$0.0015 \\
\hline
\end{tabular}
}
\end{table*}

\begin{table*}[t]
\centering
\caption{Tracking Error Standard Deviation Comparison}
\label{tab:tracking_error_std}
\resizebox{\textwidth}{!}{%
\begin{tabular}{lcccc}
\hline
\multirow{2}{*}{\textbf{Algorithms}} & \textbf{4 AUVs Tracking 2 Targets} & \textbf{6 AUVs Tracking 2 Targets} & \textbf{8 AUVs Tracking 3 Targets} & \textbf{10 AUVs Tracking 3 Targets} \\
\cline{2-5}
 & $Mean \pm SD$ & $Mean \pm SD$ & $Mean \pm SD$ & $Mean \pm SD$ \\
\hline
VGG-MADiffRL & 0.1755$\pm$0.0000 & 0.1818$\pm$0.0000 & 0.1748$\pm$0.0002 & 0.1938$\pm$0.0001 \\
DSBM & 0.1917$\pm$0.0003 & 0.1837$\pm$0.0080 & 0.1952$\pm$0.0030 & 0.1924$\pm$0.0005 \\
MA-A3C & 0.2679$\pm$0.0014 & 0.1823$\pm$0.0034 & 0.2423$\pm$0.0009 & 0.2033$\pm$0.0007 \\
MAPPO & 0.1898$\pm$0.0001 & 0.1816$\pm$0.0002 & 0.1975$\pm$0.0002 & 0.2369$\pm$0.0001 \\
MASAC & 0.2122$\pm$0.0003 & 0.2091$\pm$0.0002 & 0.1886$\pm$0.0009 & 0.2098$\pm$0.0007 \\
MAAC & 0.1806$\pm$0.0000 & 0.1952$\pm$0.0008 & 0.2626$\pm$0.0006 & 0.3483$\pm$0.0001 \\
MATD3 & 0.2008$\pm$0.0013 & 0.2976$\pm$0.0006 & 0.1909$\pm$0.0001 & 0.2239$\pm$0.0001 \\
MADDPG & 0.1970$\pm$0.0006 & 0.1799$\pm$0.0010 & 0.2281$\pm$0.0002 & 0.2478$\pm$0.0005 \\
\hline
\end{tabular}
}
\end{table*}

\subsubsection{\textbf{System Convergence Speed}}

To evaluate the training convergence of VGG-MADiffRL, we compare its convergence performance against the two categories of methods described above across four multi-AUV ad hoc network scenarios. The convergence curves are shown in Figs.~\ref{fig6}(a)--\ref{fig6}(d).

Among the general-purpose MARL baselines, MAPPO achieves the strongest convergence, benefiting from its centralized training with Generalized Advantage Estimation (GAE) and a stochastic Actor policy that balances global coordination with policy diversity. MASAC, MAAC, MATD3, and MADDPG converge more slowly, particularly in larger-scale scenarios (8 AUVs Tracking 3 Targets and 10 AUVs Tracking 3 Targets), where increased agent interactions amplify multi-agent non-stationarity. Their deterministic or entropy-regularized policies cannot adequately model the interdependent action distributions needed for coordinated tracking under dynamic ad hoc topologies.

Among the underwater-specific methods, DSBM and MA-A3C train stably but converge to lower returns than VGG-MADiffRL. DSBM uses dynamic-switching attention for multi-target tracking; it performs moderately but plateaus early because its discrete switching mechanism restricts the expressiveness of continuous cooperative actions. MA-A3C adopts a hierarchical software-defined architecture with advantage-attention actor-critic and advantage resampling; it converges steadily but reaches a lower final return because its deterministic policy gradient limits action expressiveness and its reward-weighted attention compression discards fine-grained coordination signals needed under fast-changing ad hoc topologies.

VGG-MADiffRL converges faster than all compared methods across all scenarios. During early training, the critic value-guided mechanism drives rapid policy improvement: the diffusion policy generates actions via differentiable sampling, and a joint loss combining policy gradient objectives with Q-guidance terms from global dual-Q network outputs steers updates toward high-value regions. Batch updates that start after the replay buffer reaches a minimum size suppress small-sample bias and improve sample reuse. In later training, the algorithm remains smooth and stable. The dual-Q target networks take the minimum of two independent estimates to reduce overestimation, while soft updates avoid abrupt parameter shifts. Gradient clipping limits update magnitudes, and the inherent action continuity of diffusion-based generation helps avoid the oscillations seen in several baseline methods.

\subsubsection{\textbf{Tracking Accuracy}}
In multi-AUV ad-hoc network cooperative tracking tasks, tracking accuracy serves as a core metric for evaluating training effectiveness and policy optimization. To validate the effectiveness of the proposed algorithm, experiments are configured with AUVs initially deployed at a distance of 4.5 km from the target, and a tracking error threshold of 0.8 km is adopted for performance assessment. As summarized in Table~\ref{tab:tracking_accuracy}, the comparative results demonstrate that VGG-MADiffRL achieves the highest tracking accuracy under this scenario, significantly outperforming existing baseline methods. These results fully verify the robustness and reliability of the proposed algorithm in achieving high-precision, sustained, and stable tracking within complex, dynamic underwater ad-hoc networks.

\subsubsection{\textbf{Mean Tracking Error (MTE)}} 
In multi-AUV ad-hoc network cooperative tracking, the mean tracking error quantifies overall temporal tracking deviation and serves as a core metric for evaluating policy accuracy and stability. Under a unified experimental setup, MTE is defined as the sample mean of Euclidean distances between each agent and its target, across all timesteps and agents per evaluation episode. This formulation comprehensively reflects the error level throughout execution, rather than solely on terminal states.
Table~\ref{tab:mean_tracking_error} shows that VGG-MADiffRL achieves the lowest MTE value among all compared methods, demonstrating a more pronounced error advantage. These results indicate the proposed method enables higher-precision and more robust continuous cooperative tracking in complex dynamic underwater ad-hoc networks.

\subsubsection{\textbf{Error Standard Deviation (Error Std)}}
To characterize the stability and fluctuation of tracking errors in multi-AUV ad-hoc network cooperative tracking, this study computes the standard deviation of distance error samples between each agent and its target across all timesteps and agents per evaluation episode, defined as the Error Std metric. This indicator quantifies error dispersion: a smaller value signifies more temporally consistent tracking performance with reduced fluctuations. 
Table~\ref{tab:tracking_error_std} shows that VGG-MADiffRL achieves the lowest Error Std among all compared methods, demonstrating that the proposed algorithm not only maintains a low mean tracking error but also exhibits superior error suppression and more stable dynamic tracking performance in complex underwater ad-hoc networks. In Table~\ref{tab:tracking_error_std}, standard deviations below the reported decimal precision are denoted as $\pm 0.0000$.

\begin{figure}[bth]
	\centering
	\includegraphics[width=0.96\linewidth]{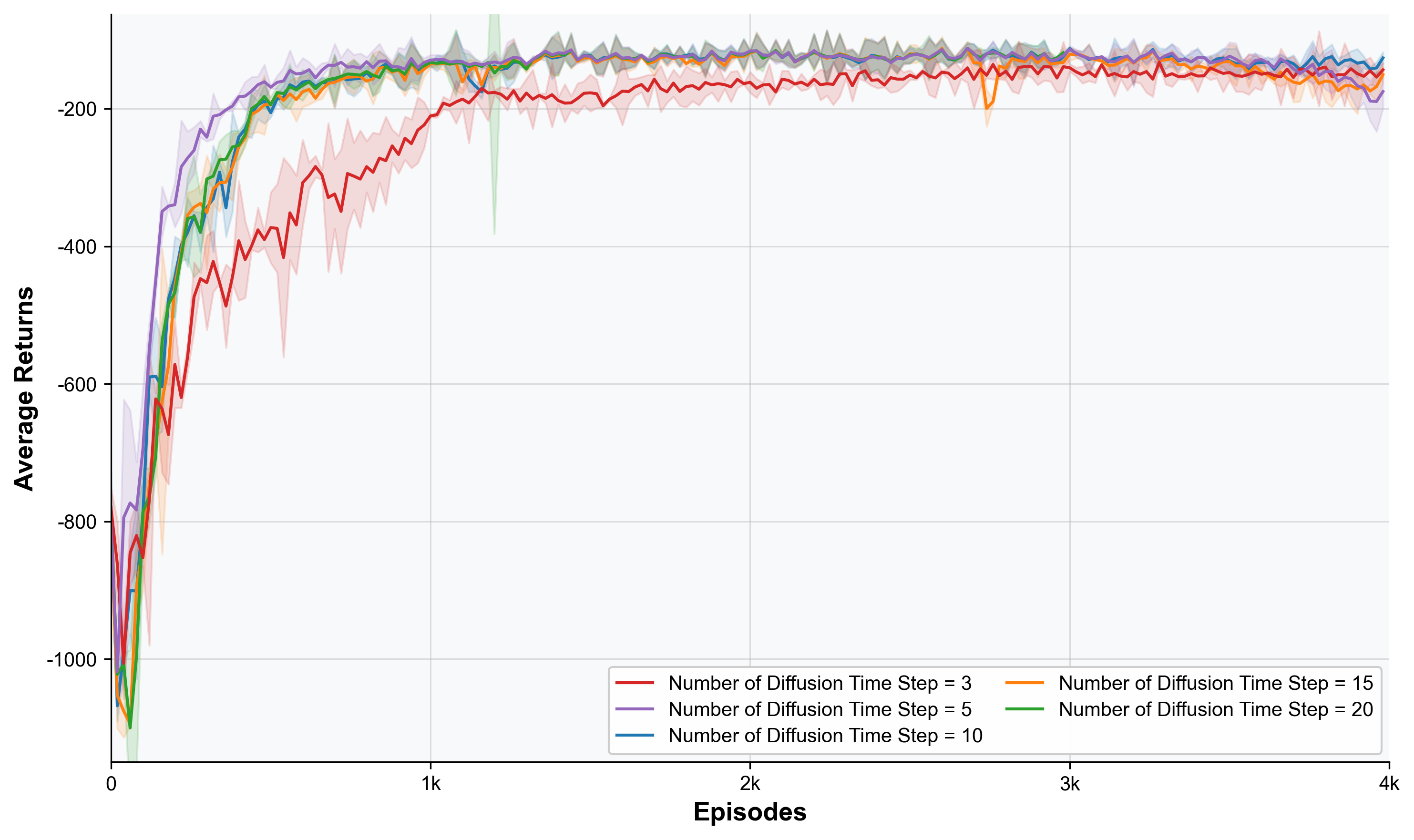}
	\caption{Convergence Speed Across Different Numbers of Diffusion Time Steps in VGG-MADiffRL}
	\label{fig4}
\end{figure}

\begin{figure*}[t!]
 	\centering
 	\subfloat[Ablation analysis in 4 AUVs Tracking 2 Targets]
 	{\includegraphics[width=0.48\textwidth]{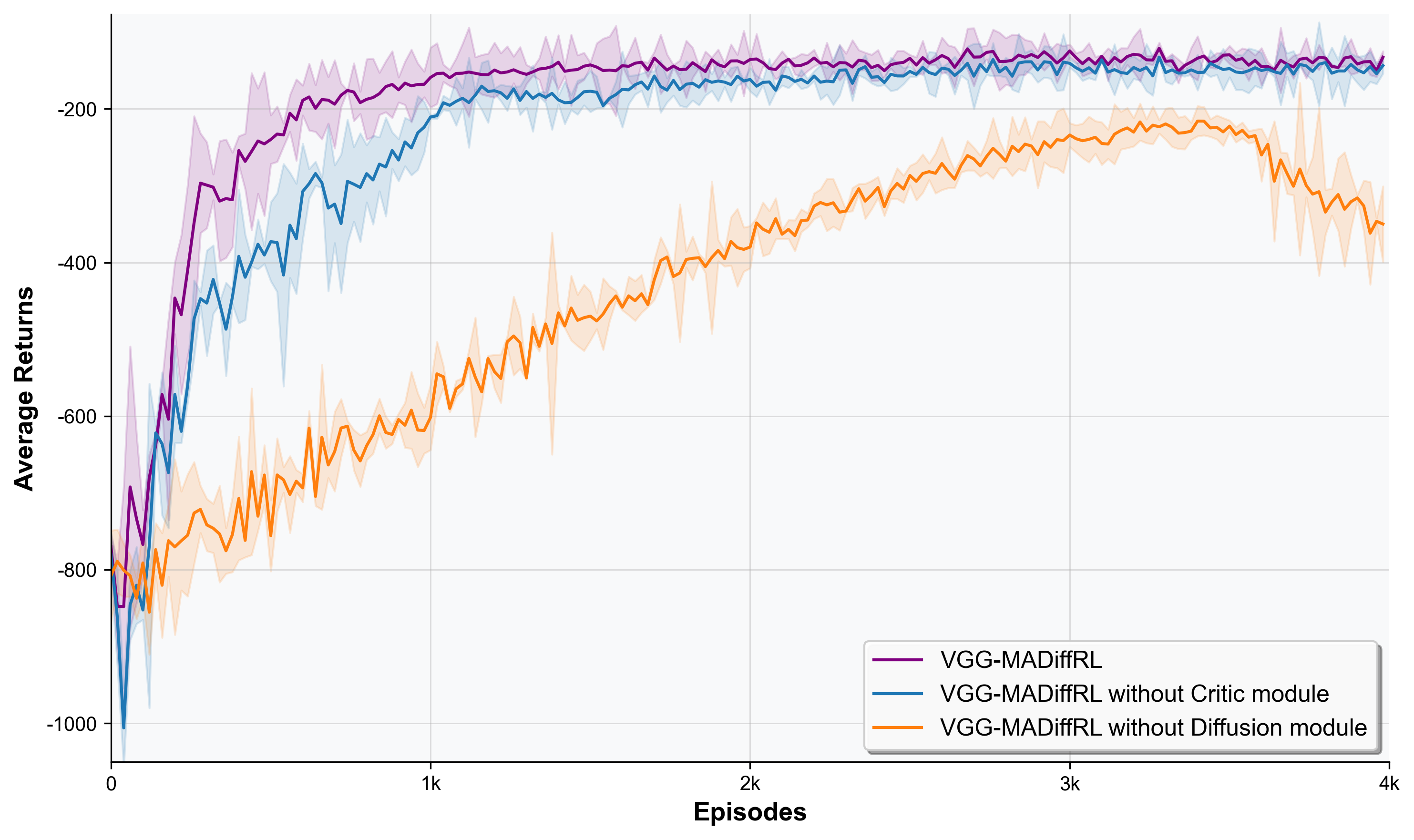}}\hfill
 	\subfloat[Ablation analysis in 8 AUVs Tracking 3 Targets]
 	{\includegraphics[width=0.48\textwidth]{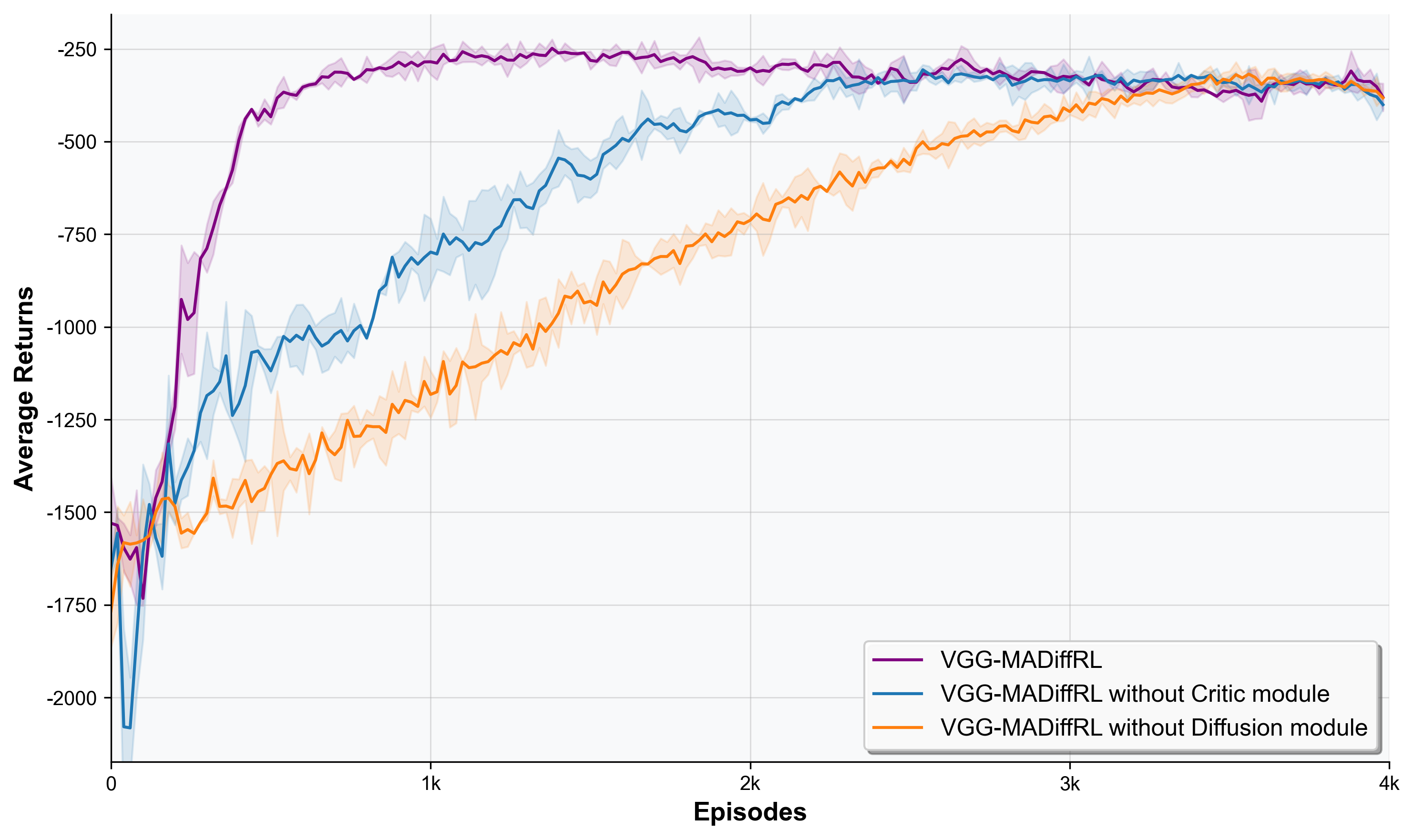}}
 	
 	\caption{Ablation Evaluation}
 	\label{fig5}
\end{figure*}

\begin{figure*}[t]
    \centering
    \scriptsize
    \begin{tabular}{@{}cc@{}}
        \includegraphics[width=0.50\textwidth]{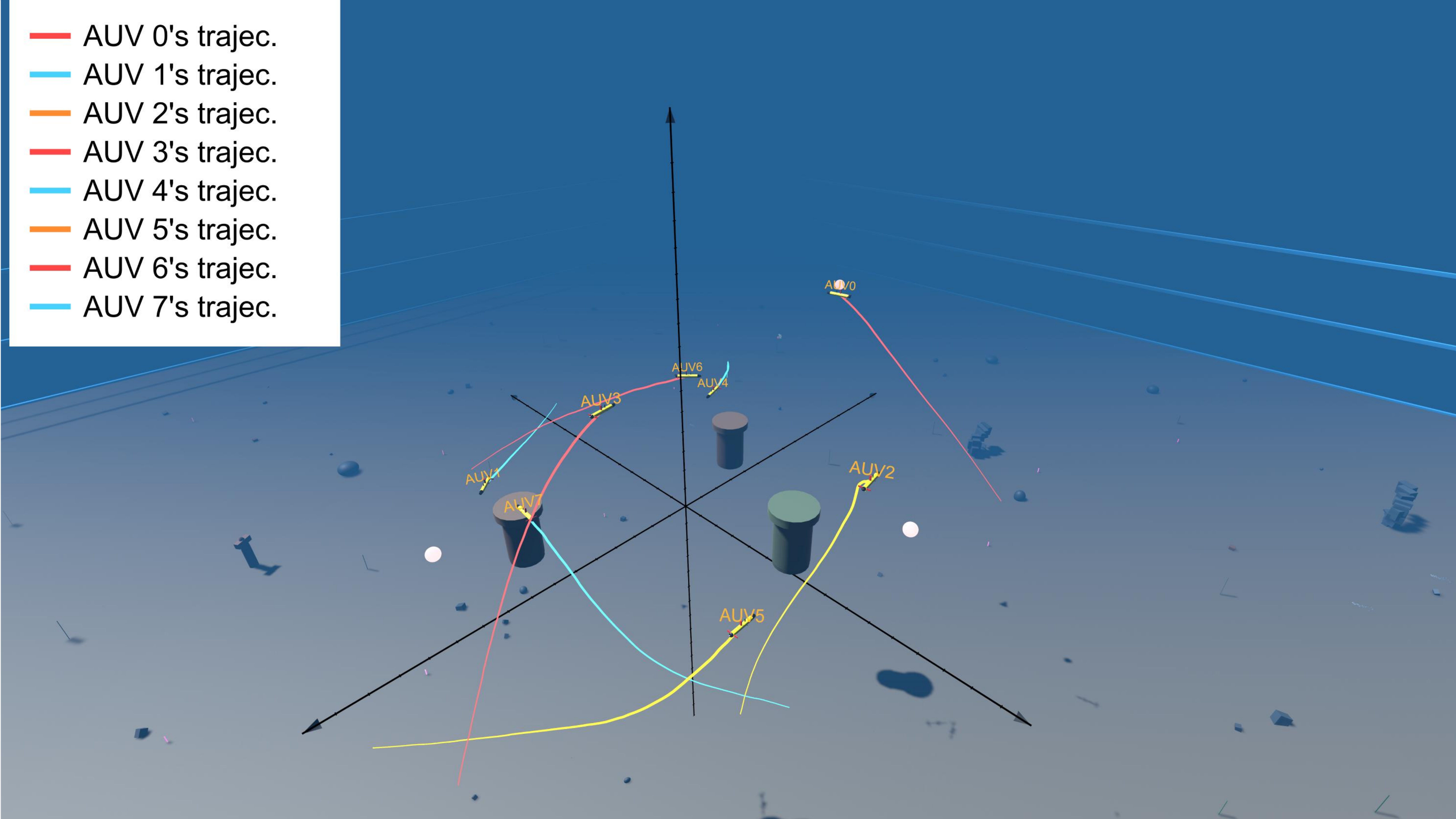} &
        \includegraphics[width=0.50\textwidth]{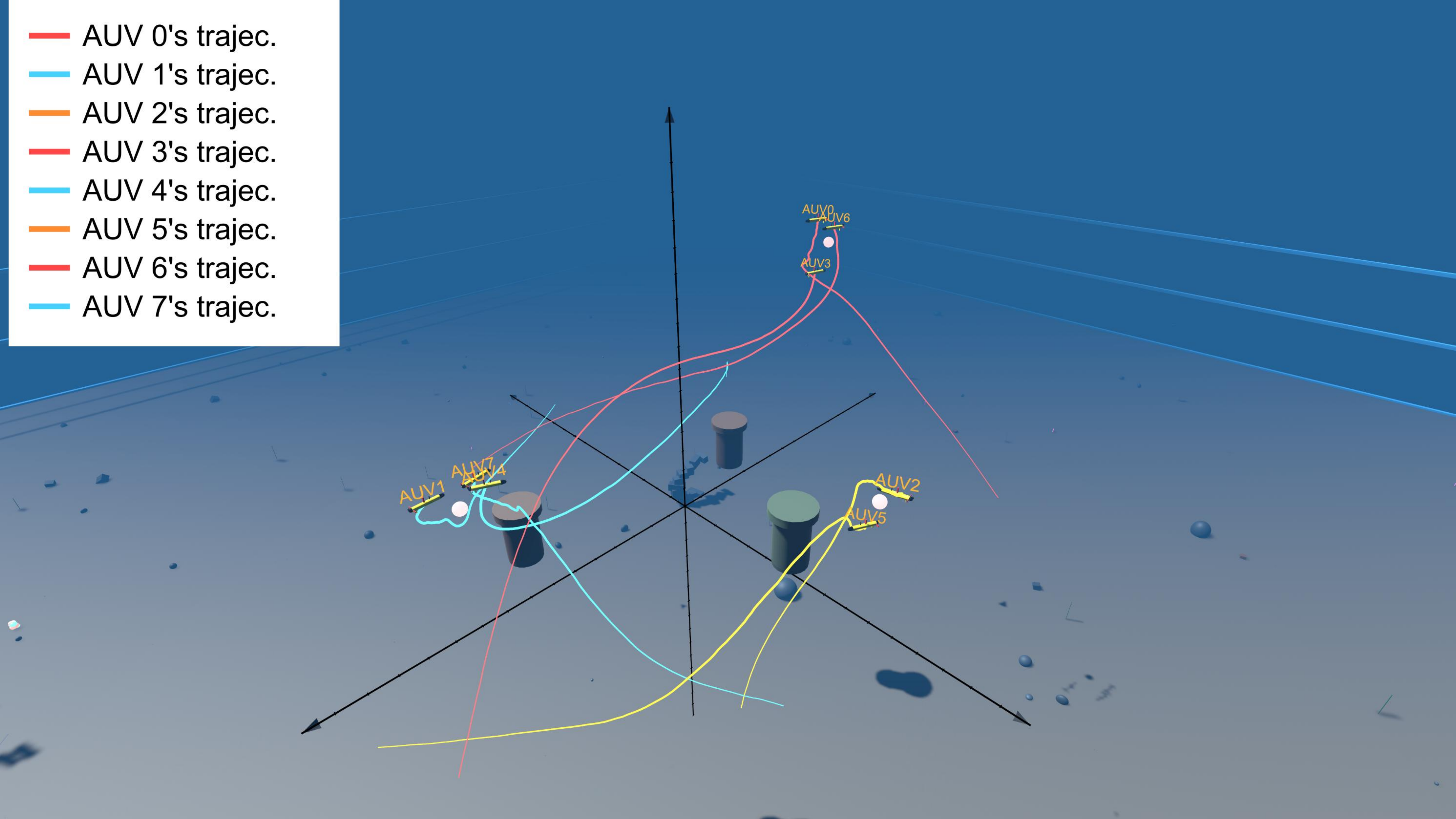} \\
        (a) Mid-Phase of 8 AUVs Tracking 3 Targets &
        (b) Final Phase of 8 AUVs Tracking 3 Targets \\[0.1em]

        \includegraphics[width=0.50\textwidth]{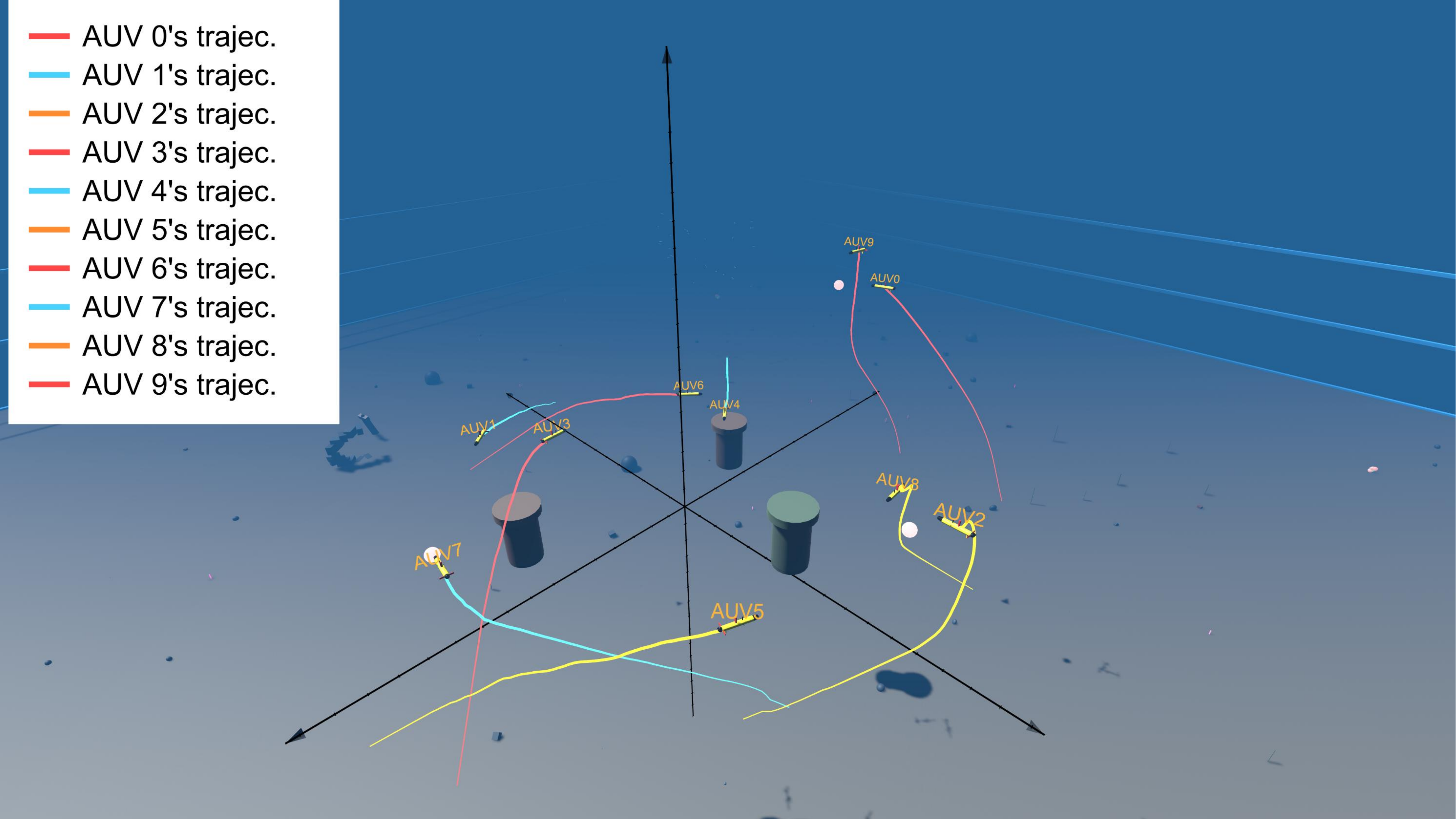} &
        \includegraphics[width=0.50\textwidth]{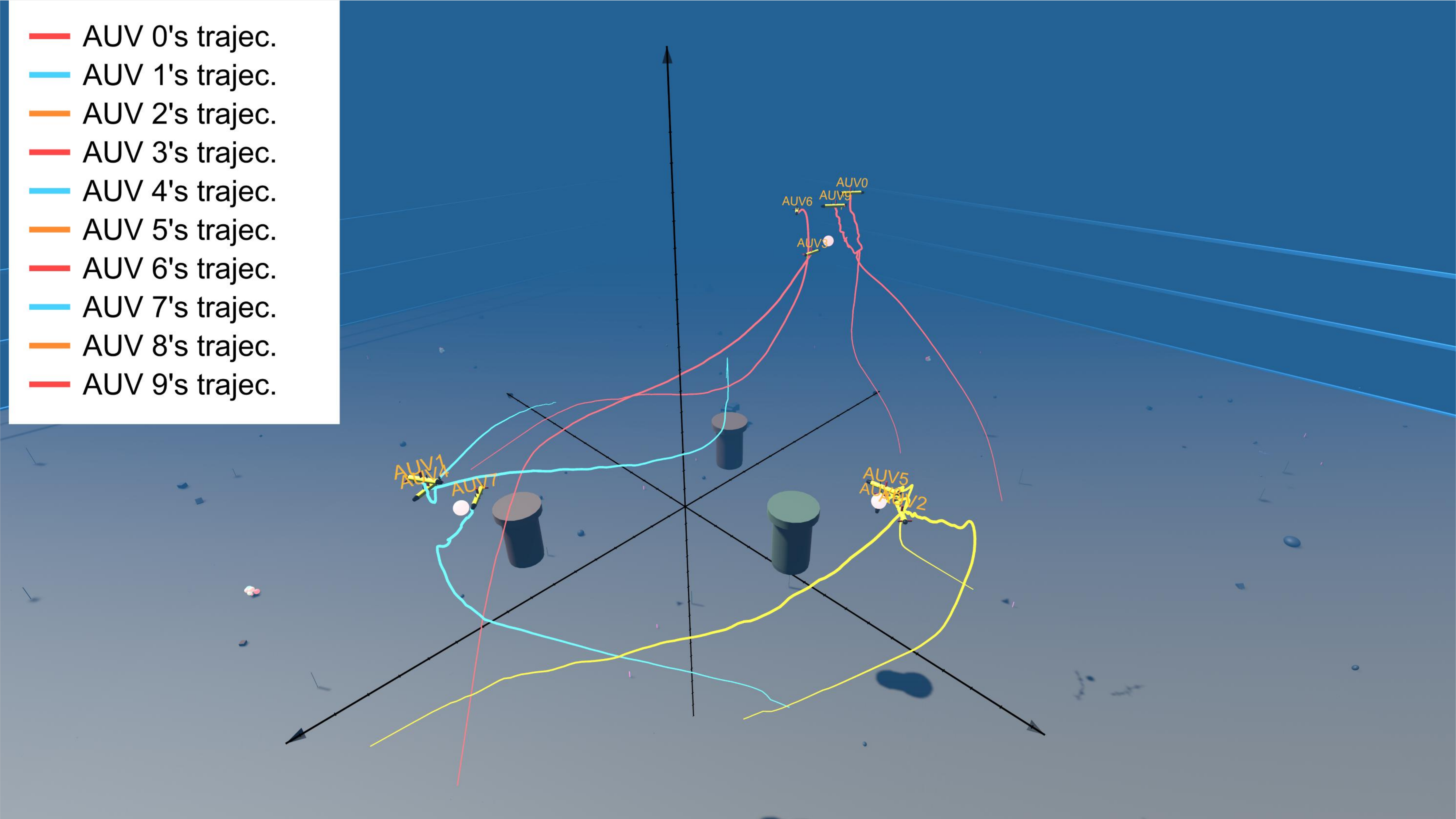} \\
        (c) Mid-Phase of 10 AUVs Tracking 3 Targets &
        (d) Final Phase of 10 AUVs Tracking 3 Targets \\
    \end{tabular}
    \caption{Availability Evaluation}
    \label{fig:multi_track_systematic}
\end{figure*}

\subsubsection{\textbf{Number of Diffusion Time Steps}}
The number of diffusion steps, the count of reverse sampling iterations in the diffusion-based policy generation process, is a critical hyperparameter affecting strategy accuracy and computational overhead. A larger number of diffusion steps enables more thorough denoising in the reverse process, theoretically yielding higher-quality action distributions and more refined policy representations, yet concurrently increases inference overhead and latency. 
Fig.~\ref{fig4} illustrates a scenario of 4 AUVs in an ad-hoc network Tracking 2 Targets, where comparative evaluations under different diffusion step settings are conducted under unified experimental conditions to analyze their trade-off effects on control precision and efficiency. Experimental results demonstrate that the adopted diffusion step configuration achieves a better balance between tracking effectiveness and computational cost, while maintaining satisfactory cooperative tracking performance in multi-AUV ad-hoc networks.

\subsubsection{\textbf{Ablation Evaluation}}
To systematically evaluate the contribution of each key module in the proposed method for multi-AUV ad-hoc networks, an ablation study is conducted. The following comparative variants are configured: (1) removal of the value-gradient-guided reverse diffusion mechanism (excluding value function guidance during action sampling); and (2) replacement of the diffusion policy module with a conventional deterministic policy network to examine the individual impact of diffusion modeling on performance. All other training configurations remain identical to ensure a fair comparison.
Fig.~\ref{fig5} shows the complete method consistently outperforms both ablated variants in convergence stability and cumulative return. These results demonstrate that both the value-gradient guidance mechanism and the diffusion-based policy module play critical roles in enhancing cooperative tracking performance, thereby validating the effectiveness of the proposed architectural design.

\subsubsection{\textbf{Availability Evaluation}}

%需要引入Ocean-gym的阐述

To show the training convergence and cooperative tracking performance of the proposed method, we build a high-fidelity underwater simulation environment using the 3D modeling and physics engine of Unity. Fig.~\ref{fig:multi_track_systematic} visualizes the cooperative tracking process and environment configuration, showing the mid and late stages of 10 AUVs tracking three targets and 8 AUVs tracking three targets. Yellow moving entities represent AUVs, glowing spheres represent dynamic targets, spirals represent obstacles, and colored trajectory lines mark the historical path of each AUV. The simulation reproduces complex underwater dynamics (acoustic communication constraints, ocean current disturbances, and time-varying network topologies) and provides a reliable platform for validating the stability and effectiveness of the algorithm. All AUV nodes, targets, and obstacles are modeled and rendered in real time, enabling direct assessment of the algorithm's operation in dynamic ad-hoc networks.

% By comparing trajectory convergence rates and path planning quality across different AUV-to-target ratios, this systematic validation confirms the engineering practicality of the proposed method for real-world marine operations.

% \begin{table*}[t]
% \centering
% \caption{Path Length Comparison of Algorithms in Multi-AUV Tracking Scenarios (km)}
% \label{tab:path_length}
% \resizebox{\textwidth}{!}{%
% \begin{tabular}{lcccc}
% \hline
% \multirow{2}{*}{\textbf{Algorithm}} & \textbf{2 AUVs / 1 Target} & \textbf{4 AUVs / 2 Targets} & \textbf{6 AUVs / 2 Targets} & \textbf{8 AUVs / 3 Targets} \\
% \cline{2-5}
%  & $\mu \pm \sigma$ & $\mu \pm \sigma$ & $\mu \pm \sigma$ & $\mu \pm \sigma$ \\
% \hline
% SD-MARL & $1.93 \pm 0.0053$ & $3.48 \pm 0.0044$ & $6.10 \pm 0.0341$ & $6.80 \pm 0.0257$ \\
% DBSM & $2.54 \pm 0.0192$ & $4.91 \pm 0.0424$ & $6.26 \pm 0.0475$ & $6.98 \pm 0.4590$ \\
% MA-A3C & $2.22 \pm 0.0385$ & $4.87 \pm 0.0641$ & $6.78 \pm 0.0405$ & $8.36 \pm 0.0686$ \\
% MAPPO & $3.34 \pm 0.0330$ & $5.68 \pm 0.0459$ & $8.55 \pm 0.1274$ & $13.09 \pm 0.0997$ \\
% MASAC & $3.33 \pm 0.0127$ & $4.39 \pm 0.0311$ & $10.80 \pm 0.0768$ & $11.60 \pm 0.0528$ \\
% MAAC & $2.62 \pm 0.0020$ & $4.48 \pm 0.0054$ & $7.20 \pm 0.0024$ & $11.20 \pm 0.0496$ \\
% MATD3 & $2.20 \pm 0.0228$ & $4.66 \pm 0.0019$ & $8.52 \pm 0.0011$ & $14.33 \pm 0.0069$ \\
% MADDPG & $3.38 \pm 0.0031$ & $5.56 \pm 0.0020$ & $11.37 \pm 0.0286$ & $14.74 \pm 0.0252$ \\
% \hline
% \end{tabular}
% }
% \end{table*}

\section{Conclusion}\label{Section:7}

This paper investigated cooperative target tracking in multi-AUV ad-hoc networks and proposed the MDCA hierarchical control architecture together with the VGG-MADiffRL algorithm. MDCA decomposes global coordination into three layers (global intelligent control, local online training, and physical action execution), enabling synergistic optimization under the CTDE paradigm. VGG-MADiffRL introduces three key innovations: a diffusion-based policy that replaces the deterministic Actor to model complex continuous action distributions; a dual-objective joint optimization mechanism that combines Q-guided and policy gradient losses to stabilize training under volatile topologies; and a value-gradient-guided reverse sampling mechanism that steers the denoising process toward high-return action regions, reducing ineffective sampling and improving policy robustness.

Extensive experiments across four multi-AUV tracking scenarios demonstrate that VGG-MADiffRL consistently outperforms seven state-of-the-art MARL algorithms in convergence speed, tracking accuracy, mean tracking error, and error stability. Ablation studies confirm that both the value-gradient guidance and the diffusion policy module contribute substantially to overall performance.

Several directions warrant further work: (1) optimizing underwater obstacle avoidance to reduce potential AUV damage; (2) balancing energy consumption among AUVs to extend system endurance; and (3) designing robust control frameworks that explicitly account for unstable underwater acoustic communication.

% \begin{enumerate}[]
% \item Optimizing underwater obstacle avoidance mechanisms for ET-AUVs to mitigate potential damage;

% \item Balancing energy consumption among the AUVs to enhance the endurance of multi-AUV cooperative tracking systems;

% \item Designing multi-AUV system robustness control framework when the unstable underwater communication (e.g., the underwater acoustic-based communication) has to been taken into account.

% \end{enumerate}

\appendices

%\section*{ACKNOWLEDGEMENTS}

\ifCLASSOPTIONcaptionsoff
  \newpage
\fi



%\bibliographystyle{IEEEtran}
\bibliographystyle{IEEEtran}
\bibliography{ref}

\vspace{-5ex}

\begin{IEEEbiography}[{\includegraphics[width=1in,height=1.25in,clip,keepaspectratio]{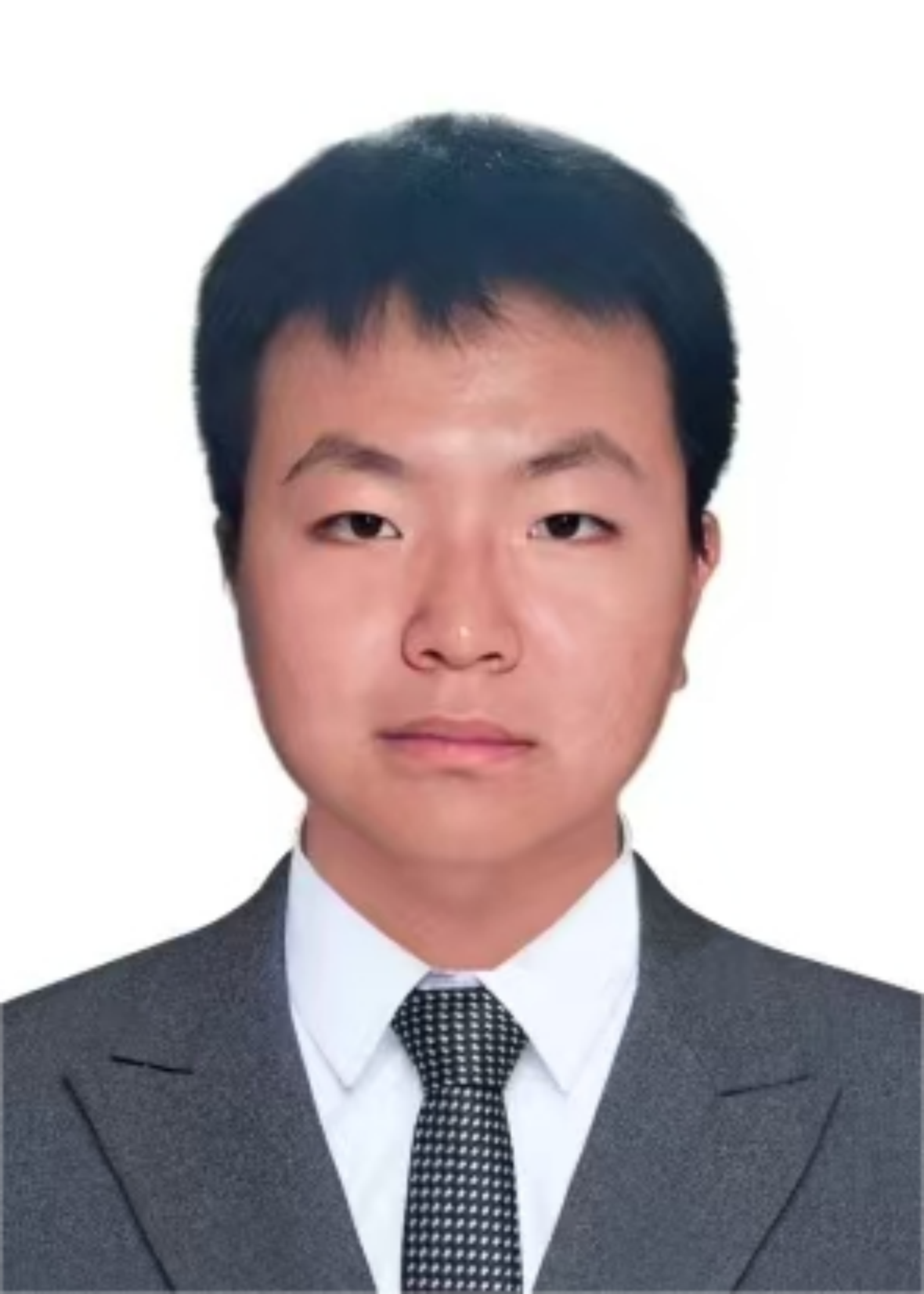}}]{Jiaao Ma}
is currently pursuing a Bachelor's degree at the Software College, Northeastern University, Shenyang, China. His research interests include reinforcement learning, diffusion models, and supervised learning.
\end{IEEEbiography}

\vspace{-5ex}
\begin{IEEEbiography}[{\includegraphics[width=1in,height=1.25in,clip,keepaspectratio]{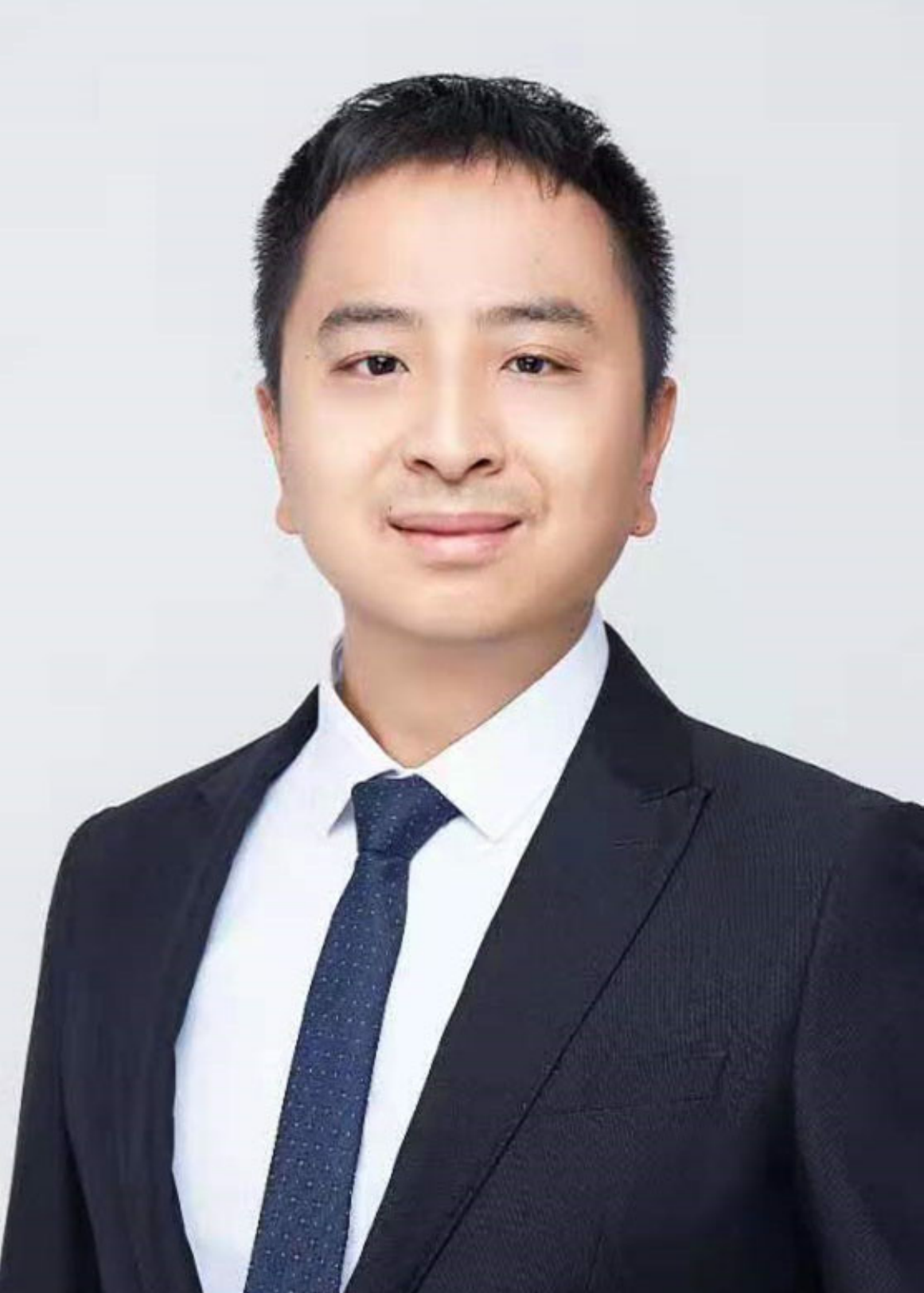}}]{Chuan Lin}
	[S'17, M'20] is currently an associate professor with the Software College, Northeastern University, Shenyang, China.
	He received the B.S. degree in Computer Science and Technology from Liaoning University, Shenyang, China in 2011, the M.S. degree in Computer Science and Technology from Northeastern University, Shenyang, China in 2013, and the Ph.D. degree in computer architecture in 2018.
	From Nov. 2018 to Nov. 2020, he was a Postdoctoral Researcher with the School of Software, Dalian University of Technology, Dalian, China.
	His research interests include UWSNs, industrial IoT, software-defined networking.
\end{IEEEbiography}
\vspace{-5ex}
\begin{IEEEbiography}[{\includegraphics[width=1in,height=1.25in,clip,keepaspectratio]{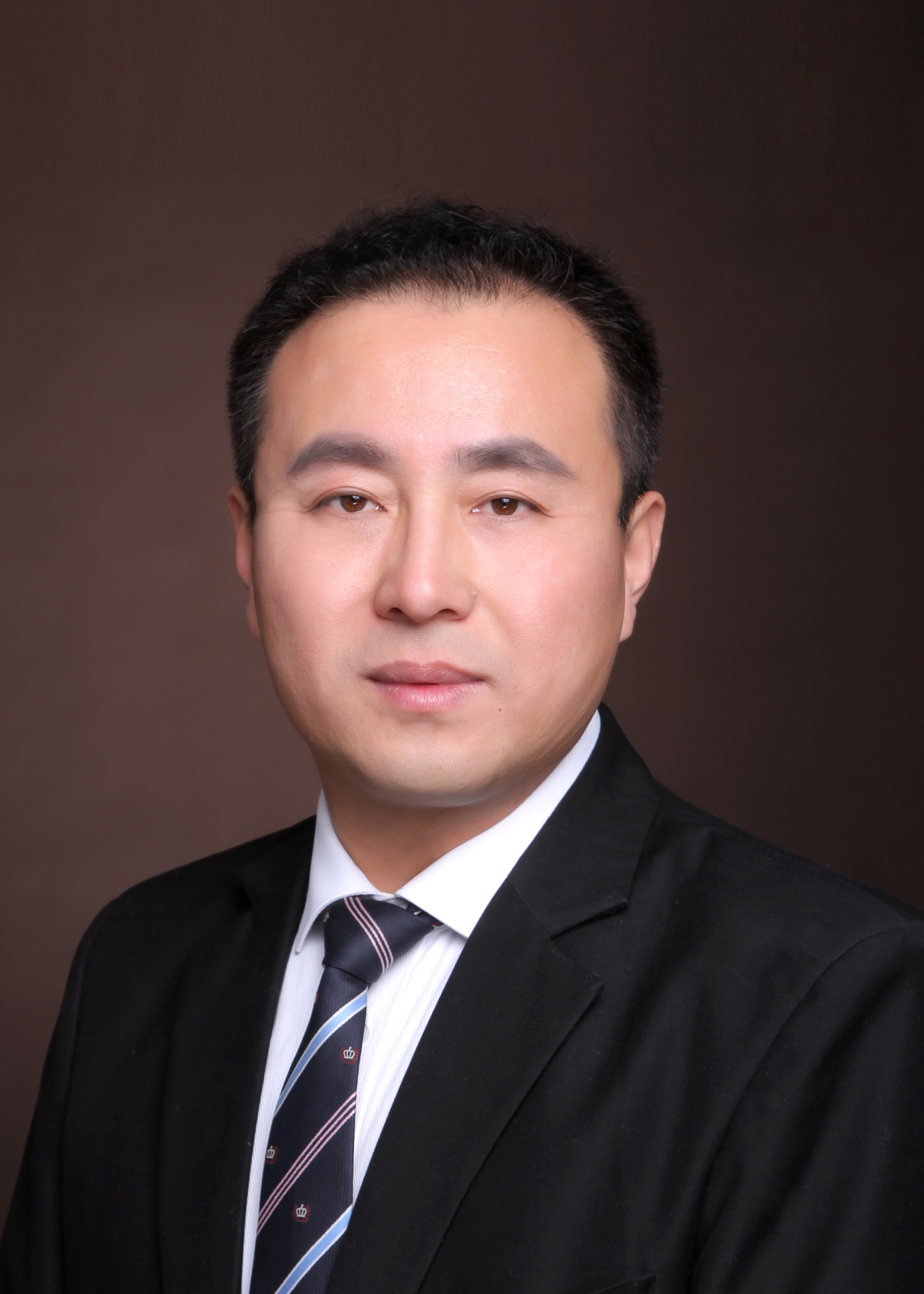}}]{Guangjie Han} (Fellow, IEEE) is currently a Professor with the Department of Internet of Things Engineering, Hohai University, Changzhou, China. He received his Ph.D. degree from Northeastern University, Shenyang, China, in 2004. In February 2008, he finished his work as a Postdoctoral Researcher with the Department of Computer Science, Chonnam National University, Gwangju, Korea. From October 2010 to October 2011, he was a Visiting Research Scholar with Osaka University, Suita, Japan. From January 2017 to February 2017, he was a Visiting Professor with City University of Hong Kong, China. From July 2017 to July 2020, he was a Distinguished Professor with Dalian University of Technology, China. His current research interests include Internet of Things, Industrial Internet, Machine Learning and Artificial Intelligence, Mobile Computing, Security and Privacy. Dr. Han has over 500 peer-reviewed journal and conference papers, in addition to 160 granted and pending patents. Currently, his H-index is 82 and i10-index is 400 in Google Citation (Google Scholar). The total citation count of his papers raises above 25000 times. Dr. Han is a Fellow of the UK Institution of Engineering and Technology (FIET). He has served on the Editorial Boards of up to 10 international journals, including the IEEE TII, IEEE TCCN, IEEE TVT, IEEE TNSM, IEEE Systems, etc. He has guest-edited several special issues in IEEE Journals and Magazines, including the IEEE JSAC, IEEE Communications, IEEE Wireless Communications, Computer Networks, etc. Dr. Han has also served as chair of organizing and technical committees in many international conferences. He has been awarded 2020 IEEE Systems Journal Annual Best Paper Award and the 2017-2019 IEEE ACCESS Outstanding Associate Editor Award. He is a Fellow of IEEE.
\end{IEEEbiography}
\vspace{-5ex}
%\clearpag

\begin{IEEEbiography}[{\includegraphics[width=1in,height=1.25in,clip,keepaspectratio]{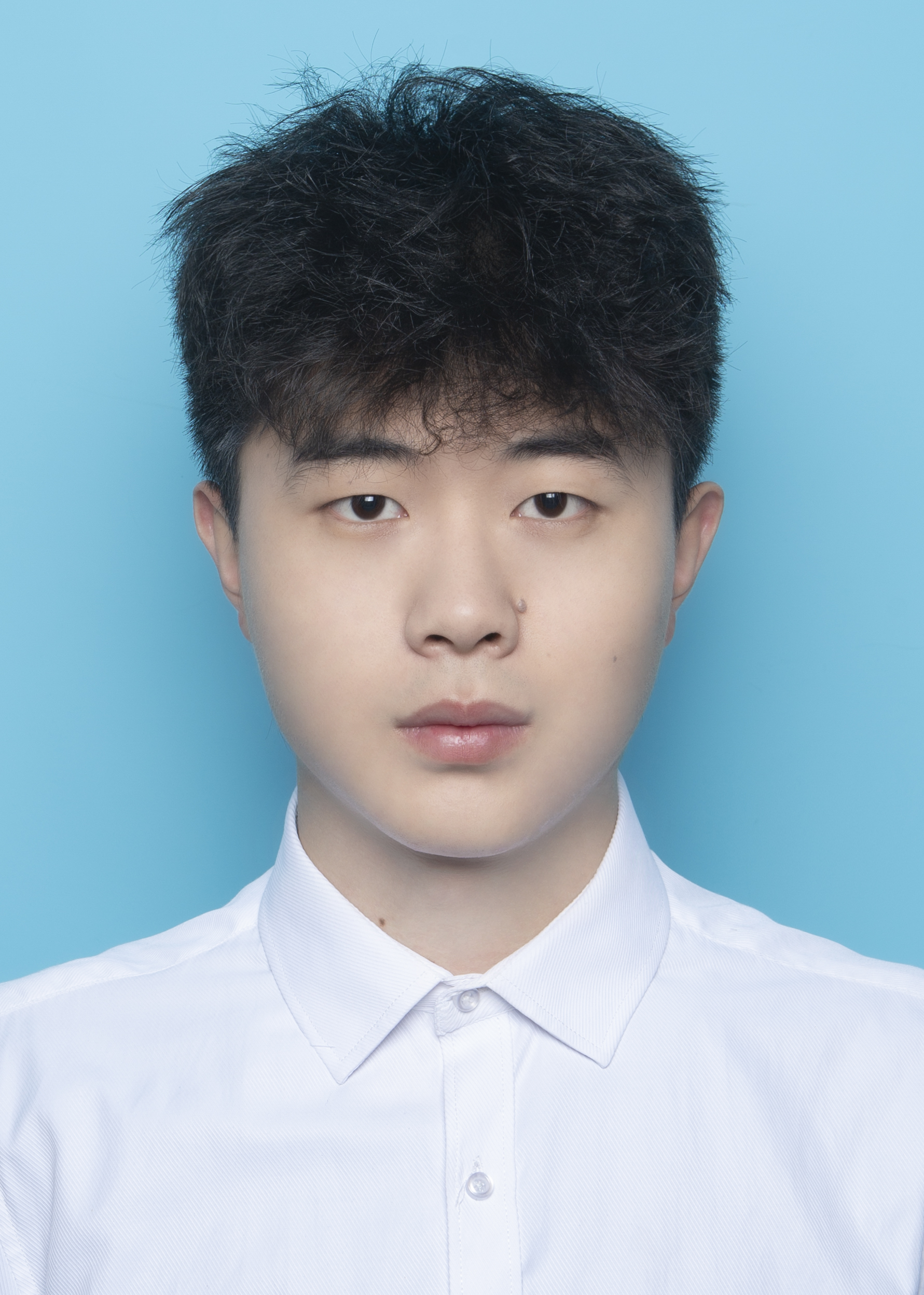}}]{Shengchao Zhu} (Student member, IEEE) received his B.S. degree in Internet of Things Engineering from Hohai University, Changzhou, China, in 2023. He is currently pursuing the Ph.D. degree with the Department of Computer Science and Technology at Hohai University, Nanjing, China. His current research interests include swarm intelligence, swarm ocean, Multi-Agent Reinforcement Learning.
\end{IEEEbiography}
\vspace{-5ex}

\begin{IEEEbiography}[{\includegraphics[width=1in,height=1.25in,clip,keepaspectratio]{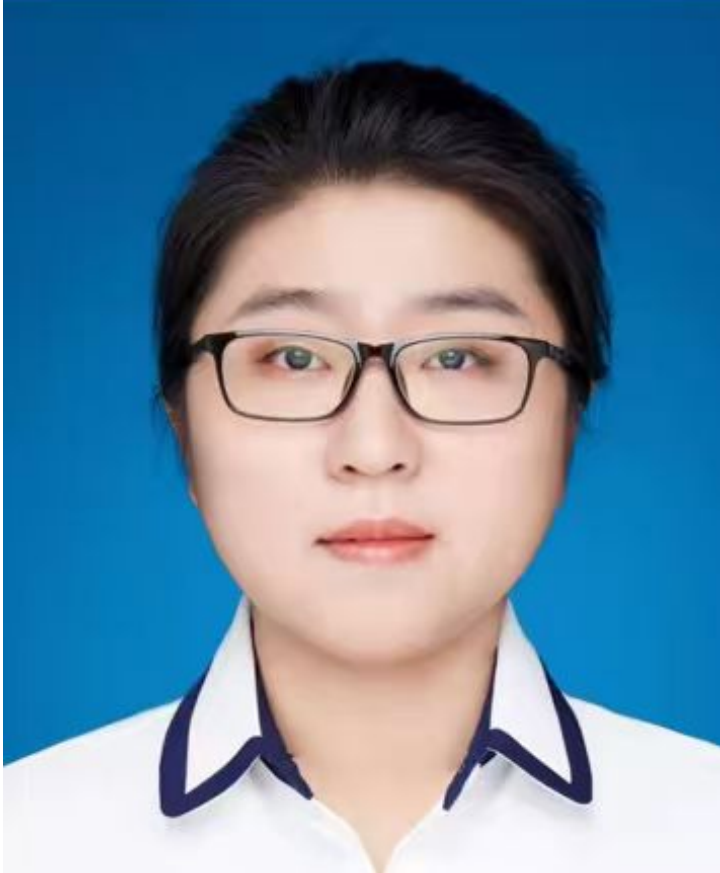}}]{Qian Zhu}
is an associate professor with the Software College, Northeastern University, Shenyang, China. She received the B.S. degree in Information and Computing Science (2006), the M.S. degree in Operation Science and Control Theory (2008), and the Ph.D. degree in Communication and Information System (2018), all from Northeastern University, Shenyang, China. Her research interests include artificial intelligence optimization algorithms, Unmanned Aerial Vehicle (UAV) technology and software development for applications.
\end{IEEEbiography}
\vspace{-5ex}

\begin{IEEEbiography}[{\includegraphics[width=1in,height=1.25in,clip,keepaspectratio]{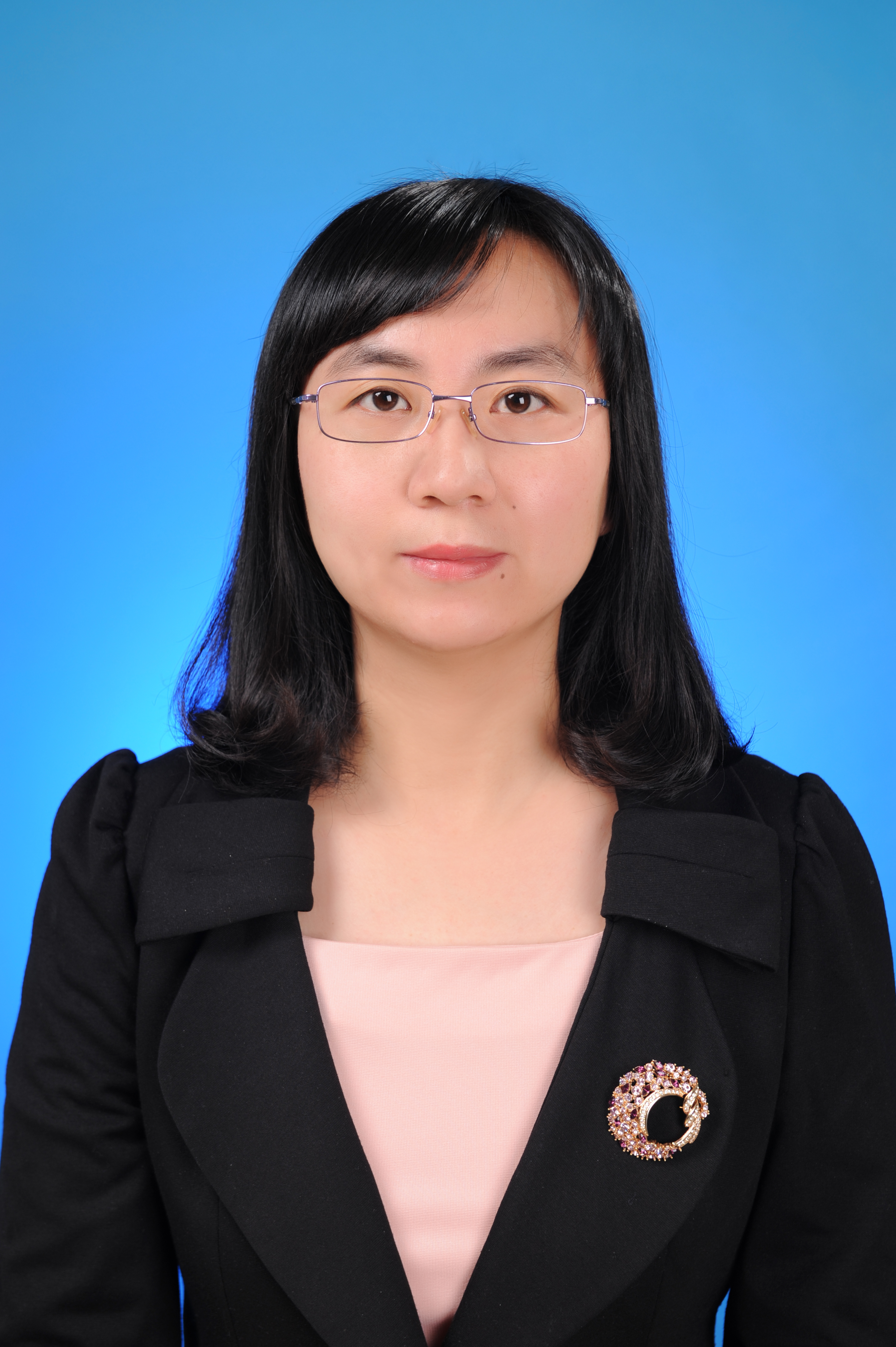}}]{Ying Liu}
received the B.S., M.S., and Ph.D. degrees from Northeastern University, Shenyang, China, in 2003, 2006, and 2012, respectively, all in computer science. She is currently an Associate Professor with the College of Software, Northeastern University. She has published over 50 articles, and refereed conference papers. Her current research interests include Service Computing and Edge Computing.
\end{IEEEbiography}

\end{document}